\documentclass[authorversion,nonacm]{acmart}
\usepackage{tabularx}
\usepackage{pdflscape}
\usepackage{adjustbox}
\usepackage{makecell}
\usepackage{graphicx}
\usepackage{subcaption}
\AtBeginDocument{%
  \providecommand\BibTeX{{%
    \normalfont B\kern-0.5em{\scshape i\kern-0.25em b}\kern-0.8em\TeX}}}

\setcopyright{acmcopyright}
\copyrightyear{2024}
\acmYear{2024}
\acmDOI{XXXXXXX.XXXXXXX}

\acmPrice{xxx}
\acmISBN{978-1-4503-XXXX-X/18/06}

\begin{document}

\title{Socially Responsible Data for Large Multilingual Language Models }

\author{Andrew Smart} 
\authornote{All authors contributed equally to this research.}
\affiliation{
\institution{Google Research}
\city{San Francisco, CA}
\country{USA}
}
\email{andrewsmart@google.com}

\author{Ben Hutchinson}
\authornotemark[1]
\affiliation{
 \institution{Google Research}
 \city{Sydney, NSW}
\country{Australia}
}
\email{benhutch@google.com}

\author{Lameck Mbangula Amugongo}
\authornotemark[1]
\affiliation{
 \institution{Namibia University of Science \& Technology}
 \city{Windhoek} 
 \country{Namibia}
}
\email{amugongol@gmail.com}

\author{Suzanne Dikker}
\authornotemark[1]
\affiliation{
 \institution{New York University}
 \city{New York} 
 \country{USA}
}

\author{Alex Zito}
\authornotemark[1]
\affiliation{
 \institution{Navanti}
\city{Washington DC} 
 \country{USA}
}

\author{Amber Ebinama}
\authornotemark[1]
\affiliation{
 \institution{Google Research}
 \city{Washington DC} 
 \country{USA}
}

\author{Zara Wudiri}
\authornotemark[1]
\affiliation{
 \institution{Navanti}
 \city{Abuja} 
\country{Nigeria}
}

\author{Ding Wang}
\authornotemark[1]
\affiliation{
 \institution{Google Research}
\city{Atlanta, GA} 
 \country{USA}
}

\author{Erin van Liemt}
\authornotemark[1]
\affiliation{
 \institution{Google Research}
 \city{Los Angeles, CA} 
 \country{USA}
}

\author{João Sedoc}
\authornotemark[1]
\affiliation{
 \institution{New York University}
 \city{New York, NY} 
 \country{USA}
}

\author{Seyi Olojo}
\authornotemark[1]
\affiliation{
 \institution{University of California, Berkeley}
 \city{Berkeley, CA} 
 \country{USA}
}

\author{Stanley Uwakwe}
\authornotemark[1]
\affiliation{
 \institution{Navanti}
 \city{Abuja} 
 \country{Nigeria}
}

\author{Edem Wornyo}
\authornotemark[1]
\affiliation{
\institution{Google Research}
\city{New York, NY} 
 \country{USA}
}

\author{Sonja Schmer-Galunder}
\authornotemark[1]
\affiliation{
 \institution{University of Florida}
 \city{Gainsville, FL} 
 \country{USA}
}

\author{Jamila Smith-Loud}
\authornotemark[1]
\affiliation{
 \institution{Google Research}
 \city{San Francisco, CA} 
 \country{USA}
}
\renewcommand{\shortauthors}{Smart \emph{et al.}}

\begin{abstract}
  Large Language Models (LLMs) have rapidly increased in size and apparent capabilities in the last three years, but their training data is largely English text. There is growing interest in multilingual LLMs, and various efforts are striving for models to accommodate languages of communities outside of the Global North\footnote{The Global North is a group of countries that are usually economically advanced and located in North America, Europe, Australia, and parts of Asia. These countries are the richest in the world according to metrics like the gross national product (GNP) per capita. Though we use the term in this paper, we acknowledge the problematic nature in its simplistic metrics-oriented definition}, which include many languages that have been historically underrepresented in digital realms. These languages have been coined as “low resource languages” or “long-tail languages”, and LLMs performance on these languages is generally poor. While expanding the use of LLMs to more languages may bring many potential benefits, such as assisting cross-community communication and language preservation, great care must be taken to ensure that data collection on these languages is not extractive and that it does not reproduce exploitative practices of the past. Collecting data from languages spoken by previously colonized people, indigenous people, and non-Western languages raises many complex sociopolitical and ethical questions, e.g., around consent, cultural safety, and data sovereignty. Furthermore, linguistic complexity and cultural nuances are often lost in LLMs. This position paper builds on recent scholarship, and our own work, and outlines several relevant social, cultural, and ethical considerations and potential ways to mitigate them through qualitative research, community partnerships, and participatory design approaches. We provide twelve recommendations for consideration when collecting language data on underrepresented language communities outside of the Global North. 
\end{abstract}

\maketitle

\section{Introduction}
Research in NLP has long grappled with modelling languages other than English \cite{ marivate2020investigating}, and NLP for languages that have been historically underrepresented on the internet since its inception is a critical challenge for AI practitioners and researchers. 
Despite this, the vast majority of the world's languages are underrepresented or not represented at all in LLMs---which have become the dominant approach to NLP and AI \cite{Amugongo2018, kirk2023understanding}. 
Recently, there are increasing efforts to make current Large Language Models (LLMs) multilingual for Natural Language Processing (NLP) tasks and include more of the world's roughly 7,000 languages \cite{pratap2023scaling, ackerman2023automatic}.
Recent large multilingual models such as mBERT and variants have capabilities---to varying degrees---in around 100 languages \cite{ruder2022statemultilingualai}.

The advent of the transformer architecture, along with the expansion of large datasets containing trillions of tokens, and the increasing computing power of technology firms, has rapidly increased the success of LLMs on primarily English-language benchmarks ~\cite{vaswani2017attention, liu2023summary}.
A debate has emerged about LLMs' ability to flexibly generalize beyond their largely English training data, and whether their mimicking of human linguistic behaviour might indicate any kind of linguistic or cognitive competence \cite{Milliere2024API}.
However, does this generalization apply outside of the dominant languages? 
The crucial issue for language inclusion and machine translation is that no one knows how to formally define what constitutes a "good" translation or whether a particular model has successfully modeled a new language \cite{poibeau2017machine}.

The amount of language data captured on devices or written on the internet has exploded, and feeds LLMs' ravenous appetites for data. 
Large language models require vast amounts of text data that is 'tokenised', good-quality machine learning models for translation, speech-to-text and other NLP tasks require additional large amounts of labelled data \cite{pratap2023scaling}.  
However the lack of sufficient speech and text data from almost all of these languages presents a major challenge to developing systems that work well outside of the major languages of English, Mandarin, Spanish, Hindi or Arabic. 
While English is a first language for only about 5\% of the world's population, English represents 44.9\% of content on the Internet \cite{johnson2022ghost}, and 93\% of the training data for GPT-3 is English \cite{brown2020language}.
Importantly, the vast majority of the world’s people do not speak English.
Mandarin Chinese (1120M), Hindi (600M), Spanish (543M) and Standard Arabic (274M) are the next largest languages \cite{dalby2003language, eberhard2019summary}. Outside these regionally-dominant languages, there are thousands of languages with few resources for creating NLP technologies. These are often described by the NLP community as "long tail" or "low resource" languages, although framings like these have been characterised by critics as Eurocentric and embodying a colonial worldview \cite{bird2022local}. 
This framing of language data as commodified resources (to be mined and refined by data "pipelines") reflects how, in the current world of massively data-hungry computational systems, data is now economically more valuable than crude oil \cite{parkins2017world}.
These computational systems happen to also be energy-hungry, raising heated debates about their contribution to setting back efforts to curb climate change.
This has disproportionately impacted---and will continue to disproportionately impact---vulnerable communities, which often are the same communities that are digitally underrepresented~\cite{strubell2019energy, ensmenger2021cloud,hagerty2023climate_divide}. 

While many have pointed to the need for more representative language data sets to improve access to LLMs and other AI tools, we stress the importance of socially responsible and ethical approaches to building those data sets.
The increased interest in data---including from languages of previously colonized people, Indigenous people, non-Western, or minoritised languages---raises many complex sociopolitical and ethical questions around algorithmic colonization \cite{birhane2020algorithmic}, consent \cite{bird2020decolonising}, ethnic identity \cite{noonan2006rise}, linguistic rights \cite{skutnabb2006linguistic}, cultural safety \cite{amugongo2023invigorating}, and data sovereignty \cite{theodore2023maori}.
While the new generation of machine learning-based language models have the potential to assist in cross-community communication and translation, language preservation, cultural transfer, and could help with economic development; this paper takes the position that the the project of data collection and massively multilingual capabilities in LLMs cannot be divorced from the cultural, political, and historical embeddedness of these languages.
In other words, language data is not just data. These models and their training data sets are by and large created and deployed by organizations with vast wealth and political power, and there may be few actual speakers of the languages being modeled at these organizations.
We argue further that these issues must not only be acknowledged but centered and prioritized in any approach to developing responsible multilingual language models.
We wish to avoid language inclusion in multilingual models as being part “of a larger repertoire of ‘happy talk,’ which involves a willingness to acknowledge and even revel in cultural difference without seriously challenging ongoing structural inequality” \cite{benjamin2023race} as quoted in \cite{hoffmann2021terms}.

A central dogma of mainstream NLP is that insights from linguistics and social science are not necessary for the successful solving of real-world language tasks through the application of statistical methods to large amounts of text data \cite{rao2019natural}. 
But the glut of data has arguably only intensified the need for comprehensive, thick description \cite{geertz1973thick}, contextualized, ethnographic, and qualitative understandings of the local contexts and global system dynamics at play: more data has not meant more equal representation of language uses, topics, and social, geographic and demographic perspectives---to the contrary. 
Such an approach should include a conscious effort to first gain insight into the viewpoints and mental models of those whose languages are at the center of data collection discourse. 
We argue that addressing language inclusion in LLMs requires much more focus on social and cultural understanding than on technical solutions. 

In light of these concerns, we make 5 main contributions specifically for NLP and language technologists:
\begin{itemize}
    \item We make explicit the implicit or latent theoretical assumptions that technologists have about languages and language communities and subject them to critical examination in relation to multilingual models.
    \item We briefly highlight the structural relationship between colonialism, globalization and languages in geographies outside the Global North and West, and we suggest that a historically grounded structural understanding of language communities can assist in responsible and more accurate, representative technology development that meets the needs of actual users globally.
    \item We outline how non-Western cultures, epistemologies and ethics present a challenge to the way in which LLMs prioritize a specific point of view, while silencing or ignoring alternative ways to categorize and represent.
    \item We highlight Indigenous projects and efforts in NLP and language modelling from people outside of the US and Europe, for example African NLP work. 
    \item We list 12 recommendations for changing the data practices of multilingual language model development based on the foregoing research and discussion.
\end{itemize}
\section{challenges for socially responsible multilingual data}
\label{sec:challenges}
LLMs with capabilities in dozens or hundreds of languages require enormous amounts of multilingual data, and model developers are unlikely to have competence in most of the languages represented in the training data. In this section we identify six challenges for model developers who wish to take a socially responsible approach to creating and curating the training datasets. These challenges all stem from the tendency of dominant methodologies in LLM development to consider language data as the primary objects of consideration---indeed, to ideologize language \emph{as} data---thus abstracting, overlooking and disenfranchising the communities who use, maintain, shape and govern those languages. In the process, technologists decontextualise languages and their technologies from their cultural, historical and societal contexts, minimalising consideration of cultural interpretation, power relations, and colonial histories of subordination and abstraction.

\subsection{Language ideologies of Natural Language Processing}
\label{sec:Language ideologies}

Language technologists tend to conceptualize a language community as a derived social entity defined by its ability to communicate fluently in the language, which is abstracted and prior in the latent theory. However, defining a language and relating it to its community are not trivial \cite{patrick2004speech}. The language community which is rarely discussed language technology discourse, despite being the more important locus of our ethical responsibilities.

Language technologists use the phrase "natural languages" to distinguish languages that occur "naturally" in human societies, in contrast with "artificial languages" which include computer programming languages. Western technologists inherit a language ideology rooted in 19th-century European nationalism, during which languages had increasing associations with nation-states and emerging national identities \cite{lupke2015ideologies,burke2004languages,leonard2017producing}. Languages and dialects are assumed to have assigned and agreed names \cite{kramer2022language}, with names frequently linked to administrative regions.  Traditional Western philology posits phylogenetic relationships with other languages, with related languages sharing descent from a common protolanguage ancestor. Our Western concept of language itself was invented in Europe, and languages are assumed to be discreet and countable \cite{gal2006migration}. Languages are assumed to have clear boundaries and be discernible from other varieties. Written language is often ``graphocentrically'' taken to be the standard form \cite{blommaert2004writing}. Language technologists have then built on this shared ideology in order to develop language naming standards such as ISO 639-3---indeed, ``what counts as a language... is about who has a three-letter code'' \cite{dobrin2009practical}---and "Language Identification" technologies which, akin to forensic fingerprinting, take a text and categorize it by returning a language label \cite{jauhiainen2019automatic}. Technologists take (digital encodings of) written text to be the default form of language, with (recordings of) spoken language called out as an edge case, rather than acknowledging the importance of writing as a foundational language technology \cite{sproat2010language}. Language technologists do not always specify the language(s) that technology can process; the expectation is that the technology works for English \cite{bender2019benderrule,ducel2022we} (and furthermore specifically US or UK English). 

We find each of these assumptions that technologists make is challenged when confronted with the multilingual realities and competing language ideologies in the Global South. Languages may not be pre-existing, but rather imagined, constructed and perceived through human activities \cite{kramer2022language}. They need not have agreed boundaries or names, and even when they do those names might not have agreed spellings. Indigenous communities may view their language as coming from, or belonging to, the land \cite{chiblow2022language}. Contact between language communities gives rise to creoles, pidgins, and dialects which blend features of two or more proximal languages. Variation may be continuous and multi-dimensional, rather than discrete, and individual texts may not be associated with a single language without additional context. The spoken form is usually the primary form, and its rich technological challenges are often overlooked \cite{chrupala2023putting}. A language may not have a written form  ("exographia"), and when they do they may have multiple competing written forms (``digraphia''),  often influenced by histories of religious expansions \cite{lupke2011orthography}. Spelling norms may exist to only varying degrees and may reproduce local language ideologies \cite{keegan2017maori}. Not every language is in the ISO language standard, and not everyone is happy about how their language is encoded in that standard \cite{morey2013language}. English is not the global default, and US and UK dialects are not the global norm. Some languages may be locally conceptualised and defined by their relationship to (rather than being prior to) a social community \cite{kramer2022language, patrick2004speech}. There may be local understandings of language ownership \cite{holton2022indigenous}, which include the language as whole as well as all of uses. Communities and individuals may have strong affiliations to languages, even in cases where a language is critically endangered and no one speaks it fluently.


\subsection{Structural issues}
\label{sec:Structural Issues}
In addition to being the dominant language represented in the training data for LLMs, English is also the primary colonial language globally, with contaminant histories of domination and subjugation of many peoples---including the suppression and elimination of Indigenous languages \cite{meighan2021decolonizing}. This has been termed "linguistic imperialism" or "linguistic hegemony" which refers to the hierarchy of languages, and addresses why some languages are more dominant than others, and it identifies what structures and ideologies facilitate this process \cite{zeng2023english, eriksen1992linguistic}. NLP and AI are historically continuous and embedded in these structural hegemonic processes and should maintain self-awareness about their own positionality in this linguistic hierarchy, indeed as is evidenced by the fact that NLP research is mainly directed toward the handful of languages dominating the internet \cite{poibeau2017machine}. Linguistic imperialism is formed by the historical, structural and cultural inequality between English and other languages, which results from structural inequality that refers to inequality related to material wealth, while cultural inequality refers to non-material or ideological inequality. The relationship between structural and cultural (or symbolic) inequality has been theorized by Bourdieu as mutually reinforcing such that symbolic capital in the form of linguistic legitimacy (e.g., the proper way of speaking, or what language to speak) serves the material interests of the dominant groups \cite{bourdieu1990logic, davis2021algorithmic, kapania2023hunt}. Linguistic hegemony is a form of what Bourdieu called symbolic violence, which refers to the ability of a dominant group to impose it symbols upon others not through physical violence but through cultural domination, the control of ideas, images, standards, icons \cite{bourdieu1990logic,39248288bedc4e18af990501e6b9e285}.

\subsection{Neocolonialism and disenfranchisement}
Questions of why and how certain languages became and remain "low-resource" tie back to histories of colonial encounter, violence, hegemony, marginalization, and resistance, and the resulting power relations experienced by their respective language communities; the current linguistic and political situation in many countries still bears both the collective generational memory and the structural traces of colonization ~\cite{birhane2020algorithmic, bright2022stability, tadei2013extractive}. Responsibly modelling these languages requires engagement with the people whose languages are being collected and understanding their unique historical and local conditions. For example, English,  French, or another European or colonial language may be the official language of the country and thus the primary or only gateway to formal power structures including but not limited to legislature, judiciary, formal economies, education systems, donor/aid structures but this does not indicate that the majority of the population use or have access to these colonial languages. Instead, the official languages are reserved for the more educated and socially elite populace while other widely spoken languages that are Indigenous to the colonized region have been forced to take a lesser role in formal communication thus, further limiting the opportunities for low-resource languages to attain more digital representation. 

Furthermore, arbitrary country borders drawn by colonizers intentionally separated organic language communities - for example, the line dividing Hausa communities between Anglophone Nigeria and Francophone Niger was not created with respect to homogeneous language, ethnicity, culture and tradition in mind. Along with the brutal severance of language communities that once were united, came the coerced union of people that speak divergent languages and live with drastically varying historical and cultural codes. This ultimately led to the strategic over-reliance on the colonial language amongst the Indigenous population because the arbitrary borders created a rigged, catastrophic issue of synchronizing across hundreds of distinct languages and cultures that never coexisted before the colonial invasion\cite{Faloyin2022}. 

\subsubsection{Disenfranchisement}
Disenfranchisement is to deprive people of civil privileges, rights of citizenship or constitutionally protected rights, and especially the right to vote \cite{39248288bedc4e18af990501e6b9e285}. Historically, colonizing powers actively deny colonial subjects or citizens basic civil or human rights \cite{scham2001archaeology}. The colonial traces of disenfranchisement still shape life for millions of people - as recently as seventy years ago the American civil rights movement had to argue, march, protest and in many cases die to establish that Black disenfranchisement and other forms of dehumanization were grave inequities \cite{39248288bedc4e18af990501e6b9e285}. As discussed in section \ref{sec:Structural Issues}, the historically continuous level of symbolic
power that colonial languages such as English have (also known as symbolic violence), that a powerful entity has, continues to legitimize its ideals,
symbols, and ideologies and de-legitimized or destroyed those of disempowered groups. The opportunity for NLP is to help correct these historical inequities through access to technology if that access is done through reversing the power dynamic outlined here.

We include here linguistic disenfranchisement which is the denial of linguistic rights such as access to education in one's native language, or the marginalization of local languages from administrative and economic inclusion \cite{skutnabb2006linguistic}.

\subsection{Language endangerment and cultural identity}
In many places, Indigenous and other non-dominant languages are seen as “vernacular” (in a dismissive sense) compared to "esteemed" colonial languages (or even as "dialects" with a pejorative connotation that they are less than languages \cite{kramer2022language}), and many under-served languages may continue to become more marginalized in the face of globalization, until they eventually disappear ~\cite{mihas2013responses}. More than half of the world’s approximately 7,000 signed and spoken languages are currently endangered, and without intervention, language extinction is a real threat for many of them. As NLP and LLMs advance, there is an opportunity to assist communities not only with the preservation of their languages but also with empowering integration and engagement of their languages with globally connected networks - in sharp contrast to how many have been historically excluded from major global systems or relegated to their margins. But this must be done with intention and with the active involvement of local communities. Technology transfer---a transfer of resources and know-how regarding LLMs and related language-based technologies currently held by institutions like tech firms and universities, to researchers and other agents of knowledge and culture within language communities---cannot be a transitive process, if it is to be successful in enabling language communities to carry out the work of language and cultural preservation on their own terms.

As mentioned in section \ref{sec:Structural Issues}, the project of large multilingual language models must contend with the fact that many languages do not have official orthographies or written systems, and thus very little digital textual representation \cite{amugongo2023invigorating, noonan2006rise}.

The question of language and ethnic identity is also very complex and not straight-forward, in many cases there is a tendency to over-emphasise (or deny) linguistic differences for political or
administrative reasons \cite{blench2012atlas}.

Additionally, lack of access to technology means that there is a dearth of recorded speech data for many languages \cite{pratap2023scaling}. At the same time, the use of mobile phones is becoming nearly ubiquitous across the world, but internet usage and access tend to reinforce existing social hierarchies; for example by privileging men, urban residents, and more educated, higher income people \cite{bender2021dangers, johnson2022ghost}. In sum, access to technology remains highly unequal globally, and even in countries with advanced technological infrastructure, basic access to the internet can be challenging for remote Indigenous communities \cite{thomas2023measuring}.

We take the position that, in solidarity with these communities (see Section~\ref{sec:relational}), advanced technology should be made available as a technological transfer but open to change according to the self-defined needs of the community. What if what a community needs is affordable internet, font and special character support, and assistive input tools and not access to the latest version of a chatbot? If what is "offered" by the extension of LLMs into low-resource language domains amounts to behavioural "nudges" to increase the consumption of imported products, then the communities comprising these new markets are right to \textit{caveat emptor} and seek agency in the process, including the right to refuse (see below).


\subsection{Language and Culture}


\label{sec:Language and Culture}
\begin{quote}
    "Language is culture" - Anonymous, Nigeria
\end{quote}

\begin{quote}
    "Communication is culture and culture is communication" -Anthropologist Edward T.\ Hall
\end{quote}

The current practice of NLP ignores the role of culture in language, but we argue that this necessarily limits the capabilities of models, but it is also a form of epistemic laziness \cite{medina2012epistemology}. Building on our argument that the implicit theory of NLP is that language is one and the same as decontextualized data, we argue that language technologists ought to partner with cultural custodians or other experts to gain more understanding.

The role of language and culture in shaping each other is disputed in the fields of sociology, anthropology, and linguistics. It is argued that the relationship between language and culture is under-theorized in sociology and anthropology, and neglected in linguistics \cite{savci2015culture}. Several scholars have argued that language and culture are co-constitutive, that is they create each other \cite{butler1999coconstructing}.  Certain branches of linguistics attempt to tackle that which is non-language or pre-language without solidly landing on a designation  \cite{paveau2006prediscourse}. In the Sapir-Whorf debate, some linguists argue that the influence of language on culture is at best minimal and two people can have similar or different cultures regardless of whether they speak the same language \cite{mcwhorter2014hoax}. While the interaction between language and culture may be beyond this paper, the advances of LLMs will make front and center of the conversation their influence on language. The impact of LLMs on language change and the accessibility of cultural concepts has yet to be understood. As an example, studies have already noted a reflection of embedded cultural values in generated AI outputs \cite{johnson2022ghost}. Understanding the possible influences of LLMs on both language and culture will be necessary in establishing appropriate policies for how the relationship of language and cultural practices is encoded, stored, and displayed.

\subsection{Non-Western epistemologies, language, meaning and place: there is no such thing as pure data}
\begin{quote}
    “Meaning does not happen in disembodied, delocalized texts.” ~\cite{burkhart2019indigenizing} 
\end{quote}


 For some non-Western epistemologies, meaning and language are bound up in the land and respect for the land, and different aspects of the land, and should express the way that respects relationships to it. Knowledge (and thus by extension its presumed precursor, data) in Western epistemology is abstracted to a core of delocalized truth that is continuous across all localities \cite{burkhart2019indigenizing}. As a product of Western epistemology, the goal of NLP research is to find common aspects or features of languages that can describe large amounts of data in the most general terms, rejecting or ignoring parts of languages that resist easy classification \cite{rao2019natural, deloria2001american}. A salient example is tonality, which is present in about 80 \% of African languages but has received little attention in NLP \cite{ruder2022statemultilingualai}. 
 
 The transformation in NLP of language into computational representations such as vectors, tensors, graphs and trees requires the digital encoding of text which is assumed to be the default form of the language (see section \ref{sec:Language ideologies}). A fundamental aspect of NLP is the idea that semantic meaning and linguistic structure can be accurately represented in abstract high-dimensional latent mathematical spaces \cite{landauer1999latent}. In other words, at the bottom NLP relies on the statistics of word frequencies and word co-occurrences, which amounts to counting the words in a text corpora with individual words as rows and using arbitrarily defined contexts (articles or documents) as columns is isomorphic with how humans process and understand language. 
 
 However, "objective" data do not exist: the creation of data is dependent on human labour, personal data, and social behaviours that accrue over long periods, through extended networks and controversial (and inevitably subjective) taxonomies ~\cite{gitelman2013raw}. And the creation of data is itself an act of interpretation - what counts as data worthy of collecting versus not is based on pre-empirical assumptions and theoretical commitments \cite{poibeau2017machine}. This fact is underappreciated, and we argue that as the expansion of language technology to include under-served languages proceeds, questions of who is collecting the data and how they are collecting the data need to be foregrounded. AI easily extends the ‘power of normalisation’ of modern institutions, among others bureaucracy, medicine, statistics (originally, the numerical knowledge possessed by the state about its population), and surveillance, which passes now into the hands of AI corporations ~\cite{denton2021genealogy}.

 In big tech, there is a narrative that language is trapped in a local optima. However, as highlighted in recent TED talk by Lelapa AI CEO Pelonomi Moiloa, language is a living archive developed over a period of time amongst a group with shared experiences and it reflects that group' way of life ~\cite{Moiloa_2023}. Language reveals to us the variance of the human experiences and mutable aspects of life that we may assume to be universal. It is this context and varying experiences, that is currently missing in how we represent languages in technology. For example most Bantu or Khoisan languages have rich agglutinative morphology, which implies that there are many ways to say the same thing. For instance, in Oshiwambo (one of the Bantu languages spoken in Northern Namibia and Southern Angola), one can say 'Kala po nawa' or 'Likwata nawa' or 'Enda po nawa' or 'Oshi iwete' or 'Ota ndi ku mono ngula' to say goodbye. This implies that metrics that evaluate the model performance may fail to recognise these non-context-breaking structures. Another example are Khoisan languages, the written form does not always represent the way words are pronounced because of the clicks, which are used as normal consonants. Additionally, in the era where LLMs claim to support local languages, even when they do not. The concept of "little" language model is attractive as it implies that developers should leverage communities' knowledge about their language, culture and little data to create tools that preserve the cultural richness embedded in languages. 
 
 Data collection process should be evaluated regularly to ensure accurate representation of the language and culture. This can involve regular feedback from the communities and adjustments to the models and data collection process as needed.



%




Many indigenous epistemologies and ontologies do not involve abstract, delocalized textual analysis (as is used in NLP and ML) ~\cite{burkhart2019indigenizing}. Meaning is tied to specific lands with specific relations between the land and people. Narratives, humour, activity, complexity, polysemy and the primacy of relationships can pose an incommensurable set of epistemic barriers which can only be bridged through direct engagement with people: a truth uttered cannot be separated from the spatial, temporal, social, and other human contextual elements of its utterance and reduced to the words spoken. We have to examine what it actually means to model language using a set of tools that have been developed within a Western-centered historical, cultural, and political context, and as such are reflective of a particular range of worldviews and are neither universal, nor abstract, nor objective. There are often multiple layers of meaning in every word which can only be understood through context and relations. As in Ubuntu ethics, the nature of meaning is cannot be isolated from relational dynamics.

\section{Decolonizing methodologies for community engagement}
\label{sec:mitigations}
In order to address the above concerns and considerations we outline here broad approaches to changing perspectives within the AI research community seeking to expand the repertoire of languages in LLMs. We apply lessons from the decolonizing research work in science and medicine, as well as our own perspectives as practitioners \cite{smith2021decolonizing, wilkins2023community}. 
\subsection{Relational ethics and Ubuntu}
\label{sec:relational}
A growing body of algorithmic injustice has demonstrated that AI has the potential to amplify existing inequalities and biases that make up our social fabric. This has prompted calls to rethink AI ethics, which has been dominated by Western epistemology and values. Relational ethics have been proposed as alternatives to Western ethics. The latter emphasizes individualism, centering on the rights, interests, and responsibilities of individuals as autonomous entities. Relational ethics, in contrast, advocate for a shift from rationality to relationality when thinking about data, personhood and justice \cite{BIRHANE2021100205}. An example of a culture/language that adheres to relational ethics is Ubuntu, which conceptualises personhood as being linked with other people. This set of normative beliefs about humanness shapes practices to embody values such as value for human life, peaceful relations and mutual respect \cite{Gade2012}. In short, Ubuntu represents moral beliefs that define the African way of life, rather than a singular ethical concept. Historically, the focus on communal welfare and social stability sometimes restricted individual rights within collective societies, such as kinship groups \cite{Abdulrauf2021}. Despite the shift towards individualism in Africa, influenced by urbanization, capitalism, global education, and media, Ubuntu remains influential across the continent. It highlights the importance of community in decision-making processes and in shaping responses to life’s challenges. Ubuntu-based principles have been applied across various domains, including business, politics, technology, and healthcare, offering a framework for ethical conduct that emphasizes collective well-being and inclusive values.

In Computing, the word "Ubuntu" has been mainly associated with the Linux-based operating system. Beyond the tech sloganeering, Ubuntu has a rich cultural and historical value. At the heart of Ubuntu are concepts such as humanness, fairness, social justice, reciprocity, the sanctity of human life, and harmonious relations, as well as tolerance, mutual respect, and consensus-building \cite{Ujomudike2015}. Compared to Western ethics, relational ethics provides new lenses to view and rethink the design, development and use of AI systems. A recent study has suggested that Ubuntu ethics can ground the design, deployment, and use of AI systems for healthcare to help clinicians deliver equitable care \cite{amugongo2023invigorating}.

In the context of LLMs, Ubuntu ethics can help development teams approach system design in an empathetic and holistic way, thereby promoting the well-being of stakeholders, especially those previously disadvantaged. Ubuntu ethics can enable this sort of holistic system design through representative dataset curation, involving all stakeholders and the communities that the LLMs intend to serve. Moreover, relational ethics such as Ubuntu will not only shape how data used to train LLMs are gathered but also data governance. As highlighted by Viljoen, potential harm to society can be as a result of how data is collected and used. To avert potential harms and biases, there is a need for alternative approaches to sovereign data governance to prioritize the interest of everyone, including communities \cite{viljoen2021relational}. A new data governance approach needs to address issues of ownership as well as informed consent. Additionally, relational ethics suggest that the process of data collection should be transparent. This can help build trust with the communities whose languages are being modeled and ensure that they have a say in how their languages are represented.

Ubuntu principles can ensure that LLMs are developed and utilized in ways that are ethical and advantageous to all stakeholders. The focus on community and collective well-being in Ubuntu will guarantee that LLMs are created with the needs and interests of the entire community in mind, rather than solely individual gains. Lastly, Ubuntu relational ethics provide ideal values to anchor ethics in the general design of LLM systems, ensuring they are informed by the experiences and perspectives of all users, particularly those from marginalized and underrepresented communities who are most susceptible to algorithmic injustice. Finally, as the race for Artificial General Intelligence (AGI) rages on, we need to establish inclusive principles and accountability mechanisms that can inform proper regulations to avert negative outcomes and create AI systems that protect the interests of all humanity. Therefore, ensures LLMs promote unity and assist in providing equitable opportunities and benefits to everyone. This can only be truly achieved by prioritizing the well-being of both individuals and the community.

\subsection{Community-based research}
Language communities and language ecosytems are infrequently given first class attention in LLM discourse, compared to language data (which itself a poor cousin of modeling \cite{sambasivan2021everyone}). We urge technologists to confront their language ideologies and assumptions, in particular around language communities being simply constructed and having obvious or transparent needs. Instead, we encourage technologists to recognise that languages are social constructs shared by communities of language users \cite{holton2022indigenous}. Echoing the \emph{Ethics Statements} adopted by the Linguistic Society of America (in 2009 and 2019), we urge researchers to consider how their research affects both individual participants and the wider community, and we posit that community involvement is central to the ethical development of NLP \cite{mager2023ethical}.

Involving communities as equal partners in research, design, and implementation of tools and measures has shown to be beneficial to all stakeholders involved in the process. For example, involving the general public in science can help offset pervasive challenges in science literacy and trust \cite{charles2020community}, and several community initiatives have demonstrated that co-design and co-ownership can lead to increased adaptation and buy-in of projects and initiatives that directly impact communities. Co-design and co-ownership is prevalent across fields and have led to several proven best practices, many of which are directly relevant here. For example, trusted community members are often recruited to act as representatives for the wider community members \cite{wef23}. Being mindful of how such representatives are involved is also proven to be critical. For example, the “Co-Design Model of Care” \cite{mckercher2020beyond} proposes various pathways to ensuring that everyone is able to identify and meet shared goals, and that sufficient guidelines and structure is provided to generate the most productive and satisfying outcomes for all stakeholders. This includes being mindful of built-in power dynamics and allowing for the time and effort to build relationships and trust. Depending on the task, participants can be included in group decision-making through several forms of participatory methods (e.g., dot voting). A study by members of Masakhane has shown great example of how communities can be involved in the co-creation of well-representative datasets for low-resource languages to improve classification tasks \cite{marivate2020investigating}

\subsection{Human rights and linguistic rights} 
The human rights system was designed to protect people in the globalization process, rather than give market forces and technology development free range. Human rights, especially economic and social rights should act as correctives to the free market \cite{skutnabb2006linguistic}. Human rights extends also to linguistic rights. Over the years, AI ethics initiatives have attracted a lot of interest, resulting in several ethical frameworks and guidelines. However, existing ethical principles often lack direct engagement with existing human rights law or guidance. To overcome this challenge, it has been argued that universal human rights - like the ones outlined in the Universal Declaration of Human Rights (UDHR) can provide the much-needed foundation that can ground and guide the ethically responsible design, deployment and use of AI systems, including LLM technologies \cite{prabhakaran2022human, mcgregor2019international}. An approach grounded in human rights can help harness the potential of LLMs for communities whose languages have historically low resources on the Internet while averting potential risks. In practice, this can mean conducting Human Rights Impact Assessments for language communities whose language data is to be included in training data. Specifically conducted linguistic rights impact assessments for NLP technology has, to our knowledge, not been carried out.  

While the UDHR represents human dignity and freedom collectively defined by all regions of the world, when it comes specifically to the languages spoken around the world, many languages have not experienced these privileges which directly impact the use and availability of technology today. Language data collection for low-resource languages should take into consideration historical and current violations of universal human rights to ensure that past transgressions do not interfere with modelling and integrating low-resource languages into LLMs. When considering the rights of people who speak languages that have been historically under-represented AI practitioners - just as the UDHR was developed (but unfortunately not widely adhered to) - must take a cross-cultural and consensus-building approach to inclusively representing languages in LLMs.

\subsection{Right of refusal} 
The mere concept of rejecting or refusing technology runs against the grain of the celebrated role tech has generally occupied in the West, wedded closely to the notion of progress itself \footnote{https://afog.berkeley.edu/programs/the-refusal-conference}. Critically, however, this is the same mindset that underpins colonial and imperial expansion: technology is conflated with progress, and progress is the same as Western epistemology and rationality. But there are many communities with incommensurate values who for their own reasons might refuse participation in data collection \cite{hoffmann2021terms, schwartz2022primum}.

The UN Declaration of Indigenous Rights recognizes the right to self-government and to maintaining social and cultural institutions, and specifically, the right of Indigenous peoples to ``maintain, protect and develop'' any ``future manifestations'', including technologies, of their cultures ~\cite{gilbert2007indigenous, collins2020intersectionality}. This right seems to imply that Indigenous peoples should be involved in any language data collection projects that impact them. An important element of our stewardship of these technologies is to respect the wishes of communities who do not want to share data or have their language digitized. 

\subsection{Data sovereignty and the entities taking rightful control} 
Data sovereignty has become a central consideration because of the large demand for training data for AI systems. Extending the human rights principles to data, regions around the world where languages have been inadvertently or strategically minimized across the Internet should have the absolute authority to control and govern language data that is to be collected, annotated, and modelled for use in LLM technologies. For this to be realized, AI practitioners must understand and follow through on the importance of fostering an equitable relationship with owners of underrepresented languages and ensure that the benefits of language inclusion are tipped largely to the scales of the owners to avoid extractive data collection methods. Kotut and McCrickard examined cases wherein Indigenous communities--based on their culture--request that their language and culture not be digitized~\cite{kotutmck}. Respecting such a wish from the communities builds trust between the community and researchers. 

In tandem with the demand, the use of data from communities by corporate entities without the consent of the owners of the said data has been addressed by a few communities, notably the Māori ~\cite{theodore2023maori}. For the Māori in New Zealand, data sovereignty implies holding any entity that accesses Māori data to the standards and laws of the Māori people, but not to the laws of the colonial government. The Māori have established Te Mana Raraunga as the Māori Data Sovereignty Network ~\cite{network2016te}. The network provides advocacy for the rights of Māori and acts as a vanguard for Māori-centric data~\cite{kukutai2020indigenous}. Masakhane, an isiZulu word that means we build together, is an organization geared towards people who speak languages across Africa. Masakhane aims to lead NLP research through a pan-African collaborative effort in conjunction with outside researchers ~\cite{emezue2020lanfrica}. Masakhane emphasizes inclusive community participatory technological advances that epitomize human dignity, welfare and equity ~\cite{orife2020masakhane}. Masakhane seeks to rectify the effects of colonialism that have had detrimental effects on the usage, support and integration of African languages into advanced technologies ~\cite{orife2020masakhane, BirhanePower}.

In addition to the above, several Indigenous communities throughout the world have been working on protecting their data. In Canada, the First Nations of Canada/Inuit/Metis–in collaboration with the First Nations Information Governance Centre (FNIGC) and the Royal Commission on Aboriginal Peoples (RCAP)-- established a health data repository for the benefit of the community~\cite{caron2023partnering}. The First Nations Language Authority (FNLA) works with First Nations communities to document, revitalize, and govern their languages. They support data sovereignty initiatives, empowering communities to control access and use of their language data ~\cite{anderson1995first}. In Columbia, Ley 1381 de 2012 recognizes the rights of Indigenous communities to control their intellectual property, including language data. The Instituto Caro y Cuervo collaborates with communities to document and preserve Indigenous languages using ethical and culturally sensitive methods~\cite{cuervo2002instituto}.

The non-profit Living Tongues Institute for Endangered Languages aids Indigenous communities in preserving their languages and advocating for data ownership and control by communities which fosters ethical research and collaboration \footnote{https://livingtongues.org/}. The Indigenous Languages Technology Alliance (ILTA) is a global network that connects Indigenous communities, researchers, and technologists working on language technology projects to promote ethical frameworks for AI development that respect Indigenous data sovereignty and knowledge ownership~\cite{kuhn2020indigenous}. UNESCO's resource platform, Clearinghouse on Endangered Languages, offers best practices and guidance for protecting and preserving endangered languages and emphasizes community-based approaches and ethical considerations in language technology development~\cite{maffi2002endangered}. As the landscape of data sovereignty evolves for Indigenous and historically underrepresented languages, AI practitioners should be keenly aware and endeavor to follow the guidance of countries and organizations that seek to rightfully control and own their language data for use in technology. 


\section{Socially-motivated recommendations for Mitigating for Data Collection Projects for large multilingual language models}
Taking the above arguments into account we summarize a set of recommendations intended to bring these issues to the fore when designing NLP research and data collection. Recall that from the vantage point of the colonized, the term "research" is inextricably linked to European imperialism and colonialism \cite{smith2021decolonizing}. When engaging communities in language research for technology, we highlight the questions that Indigenous activists and researchers ask, summarized by Linda Tuhiwai Smith \cite{smith2021decolonizing, dev2023building}: Whose research is it? Who owns it? Whose interests does it serve? Who will benefit from it? Who has designed its questions and framed its scope? Who will carry it out? Who will write it up? How will its results be disseminated?

These questions apply equally to NLP research for large multilingual models of non-dominant lanaguages: whose data is it? Whose model is it? Whose interests does the technology serve? Who will benefit from the data and the model? Who has designed the data collection and model architecture? How will the technology be shared? 

We call upon the NLP and language research community to take into account the following suggestions, see Table~\ref{tab:Recs}. Please note that these suggestions are not exhaustive, and their order does not imply prioritization. By embracing these suggestions, the NLP and language research community can contribute to more equitable and culturally sensitive language data collection and research practices. If we accept that the concerns and mitigations outlined in section \ref{sec:challenges} and \ref{sec:mitigations} are worthy research goals, we propose here a framework that can guide researchers in a high-level way to address the concerns and enable the implementation or operationalise the mitigation. 

We do not suggest that each recommendation is a direct mitigation of each concern we have outlined, however we advocate for the holistic adoption of these recommendations whenever collecting or processing data from language communities outside the dominant languages. 

The community-based researcher and partnership recommendation is intended to help avoid the lack of cultural understanding, decontextualisation, and abstract logic that pervades NLP research. The recommendations are intended as a framework to guide research questions and inform ethical considerations when planning or conducting research with Indigenous communities, or working on endangered languages. A key overarching message is to remain in a position of epistemic humility and grace, with an open mind, and intention to act with the best interests of those language communities we are working with at heart. We follow recent calls in biomedical research to focus on justice as essential to forming research partnerships with communities \cite{wilkins2023community}. 
\section{Conclusion}

Modern LLM development practices treat languages as just data, with the logic that multilingual capabilities simply require massive amounts of multilingual data. However languages are also cherished expressions of social and cultural identity, and at times are "symbols of resistance or mechanisms of state control" \cite{dobrin2014dying}. Since LLM development is centered in the privileged and dominant Global North, a socially responsible approach to languages of globally marginalized cultures demands paying attention to the needs, human rights, and power dynamics of local language communities, including to histories of cultural and linguistic subordination and marginalization. To confuse internet language data with language is akin to confusing a map for the territory. Just as a map cannot be used by Western tourists to determine visa requirements, currencies and exchange rates, nor local tipping and haggling practices, so too decontextualized language data cannot be used by Western developers to determine which technology practices are  socially responsible. Instead, ways of seeing and interpreting  need local guides to avoid negligence of local linguistic customs, concerns and aspirations. One important yardstick for what data practices are socially responsible is consideration of what practices local communities consider to be socially responsible \cite{cooper2024}.

This paper reviewed social and historical aspects of languages that researchers in NLP, FAccT and AI should take into account when working toward the otherwise laudable goal of language inclusion for large multilingual language models. We presented six challenges for model developers who wish to take a socially responsible approach to creating and curating training datasets for so-called "low-resource" languages (see section \ref{sec:challenges}). These are the 1) language ideologies of technologists and NLP researchers, 2) global and local historical structural inequities often resulting from histories of colonization, 3) neocolonial dynamics that continue in the present, 4) language endangerment and cultural identity, 5) the complex interplay of language and culture and 6) the problem of modeling languages that reflect non-Western epistemologies that are seemingly incomensurable with the decontextualization, universality, generality, and abstraction that is inherent to machine learning. 

We then outlined six approaches or methodologies that can help guide NLP researchers, the FAccT community, and others working on the expansion of LLMs to include more languages who are concerned with linguistic and epistemic justice. We highlight the fact there are many existing Indigenous-led NLP data and modelling projects ongoing that researchers can leverage before embarking on a modelling effort for underserved languages. These projects are rooted in community needs already and can provide guidance or potentially even partnership. 

\section{Acknowledgments}

We wish to thank all the many people and colleagues with whom we've discussed these issues over a period of years.

\newpage

\begin{table}[ht]
  \caption{Recommendations for socially responsible language data collection for underserved language communities.}
  \label{tab:freq}
   \begin{tabular}{>{\raggedright\arraybackslash}m{\linewidth}}
   \toprule

   \sc Methodology: Human rights and linguistic rights \\ \vspace{3pt}

    \textit{Recommendation 1: Respect the Autonomy of Local Communities.} \\
    We implore researchers to respect the rights of local communities in their decisions regarding the necessity of Large Language Models (LLMs) in their language, avoiding any imposition of Western or colonial perspectives \cite{bird2020decolonising}. \\ 

    \midrule

   \sc Methodology: Community-centered research \\ \vspace{3pt}

    \textit{Recommendation 2: Engage Local Communities as Partners} \\
    Embrace ethnographic methods to engage with local communities as partners in data collection endeavours. This approach, especially when working with communities that share a common language with researchers, fosters trust and facilitates the development of beneficial technologies \cite{bird2020decolonising}. \\\vspace{3pt}

        \textit{Recommendation 3: Understand Local Technology Needs} \\
        Recognize and address local technology needs, which may include language revitalization, preservation, migration challenges, and cross-community communication \cite{cooper2024, bird2022local}. \\ \vspace{3pt}

        \textit{Recommendation 4: Support Community-Based Language Initiatives} 
        \\Extend support to existing community-based language projects such as Māori data sovereignty, Masakhane (“we build together”) and alike, which emphasize collaborative efforts \cite{bird2022local}. \\\vspace{3pt}

    \textit{Recommendation 5: Engage with Local Community Concerns} \\
    When developing bias and safety evaluation datasets, collaborate with the local community to understand and address local cultural concerns and stereotypes \cite{dev2023building}. \\ 

\midrule

   \sc Methodology: Relational ethics \\ \vspace{3pt}

    \textit{Recommendation 6: Critical Examination of Data Collection Projects} \\
    Prior to embarking on data collection initiatives, it is crucial to critically assess their potential ramifications to ensure the welfare of the affected communities. This entails an examination of power dynamics, implicit biases, and contextual factors that may introduce inequity \cite{schwartz2022primum}.\\
   \vspace{3pt}

    \textit{Recommendation 7: Acknowledge Local Power Relations} \\
    Investigate the power dynamics within the local context and how they influence access to language-based technologies, data collection, and the potential impact of LLMs on these power relations. \\ \vspace{3pt}
    
    \textit{Recommendation 8: Incorporate Intersectionality and Qualitative Approaches} \\ 
    Apply justice-oriented intersectionality theory and qualitative approaches to data collection. Consider how layers of identity affect individuals' lived experiences, power dynamics, and marginalization \cite{collins2020intersectionality, davis2021algorithmic, kapania2023hunt}.\\  \vspace{3pt}

    \textit{Recommendation 9: Value Community Empowerment} \\
    Prioritize data collection processes that bring cultural and economic value to the community \cite{cooper2024}. Acknowledge that marginalized communities may have different perspectives on language and knowledge work compared to Western researchers \cite{bird2022local}. \\ 
\midrule

   \sc Methodology: Sovereignty and control \\ \vspace{3pt}

    \textit{Recommendation 10: Transfer LLM Technology to Global South Stakeholders} \\
    Promote the transfer of LLM technology to stakeholders in the Global South. This includes giving local stakeholders agency in determining approaches and annotator inclusion \cite{bird2020decolonising}. \\ \vspace{3pt}

    \textit{Recommendation 11: Implement Data Governance} \\
    Embrace concepts of data governance throughout the research process, including data collection. Clarify ownership and decision-making, recognizing that open-source does not equate to community-owned data. \\

\midrule
\sc Methodology: Cultural interpretation \\  \vspace{3pt}
        \textit{Recommendation 12: Ensure Accurate Language Representation} \\
        Dedicate the necessary time and resources to ensure accurate representation of languages in the work.  \\

  \bottomrule
\end{tabular}
\label{tab:Recs}
\end{table}

\newpage

\section{Positionality statement}
Given that this paper directly addresses the relationship between positionality and language modelling, it is important to state author positionality as it relates to this research. We are a diverse group of academic and industry researchers from both the "Global North" and "Global South". We are working in North America, Asia, Europe and Africa. We bring perspectives and backgrounds from these geographies and combined our author team speaks over 30 languages from all over the world. We are also a multidisciplinary team representing NLP, computer science, linguistics, HCI, anthropology, African Studies, peace and conflict studies, medicine, and economic development. We believe the composition of our research team is a reflection of the values and principles we outline in this paper. 

\newpage

\bibliographystyle{ACM-Reference-Format}
\bibliography{socialMain}


\begin{thebibliography}{104}


\ifx \showCODEN    \undefined \def \showCODEN     #1{\unskip}     \fi
\ifx \showDOI      \undefined \def \showDOI       #1{#1}\fi
\ifx \showISBNx    \undefined \def \showISBNx     #1{\unskip}     \fi
\ifx \showISBNxiii \undefined \def \showISBNxiii  #1{\unskip}     \fi
\ifx \showISSN     \undefined \def \showISSN      #1{\unskip}     \fi
\ifx \showLCCN     \undefined \def \showLCCN      #1{\unskip}     \fi
\ifx \shownote     \undefined \def \shownote      #1{#1}          \fi
\ifx \showarticletitle \undefined \def \showarticletitle #1{#1}   \fi
\ifx \showURL      \undefined \def \showURL       {\relax}        \fi
\providecommand\bibfield[2]{#2}
\providecommand\bibinfo[2]{#2}
\providecommand\natexlab[1]{#1}
\providecommand\showeprint[2][]{arXiv:#2}

\bibitem[Abdulrauf(2021)]%
        {Abdulrauf2021}
\bibfield{author}{\bibinfo{person}{Lukman~Adebisi Abdulrauf}.} \bibinfo{year}{2021}\natexlab{}.
\newblock \showarticletitle{Giving ‘teeth’ to the African Union towards advancing compliance with data privacy norms}.
\newblock \bibinfo{journal}{\emph{Information \& Communications Technology Law}}  \bibinfo{volume}{30} (\bibinfo{date}{5} \bibinfo{year}{2021}), \bibinfo{pages}{87--107}.
\newblock
Issue 2.
\showISSN{1360-0834}
\urldef\tempurl%
\url{https://doi.org/10.1080/13600834.2021.1849953}
\showDOI{\tempurl}


\bibitem[Ackerman and Balyan(2023)]%
        {ackerman2023automatic}
\bibfield{author}{\bibinfo{person}{Ryan Ackerman} {and} \bibinfo{person}{Renu Balyan}.} \bibinfo{year}{2023}\natexlab{}.
\newblock \showarticletitle{Automatic Multilingual Question Generation for Health Data Using LLMs}. In \bibinfo{booktitle}{\emph{International Conference on AI-generated Content}}. Springer, \bibinfo{pages}{1--11}.
\newblock


\bibitem[Amugongo(2018)]%
        {Amugongo2018}
\bibfield{author}{\bibinfo{person}{Lameck~Mbangula Amugongo}.} \bibinfo{year}{2018}\natexlab{}.
\newblock \showarticletitle{Understanding What Africans Say}. In \bibinfo{booktitle}{\emph{Extended Abstracts of the 2018 CHI Conference on Human Factors in Computing Systems}} (, Montreal QC, Canada,) \emph{(\bibinfo{series}{CHI EA '18})}. \bibinfo{publisher}{Association for Computing Machinery}, \bibinfo{address}{New York, NY, USA}, \bibinfo{pages}{1–6}.
\newblock
\showISBNx{9781450356213}
\urldef\tempurl%
\url{https://doi.org/10.1145/3170427.3180301}
\showDOI{\tempurl}


\bibitem[Amugongo et~al\mbox{.}(2023)]%
        {amugongo2023invigorating}
\bibfield{author}{\bibinfo{person}{Lameck~Mbangula Amugongo}, \bibinfo{person}{Nicola~J Bidwell}, {and} \bibinfo{person}{Caitlin~C Corrigan}.} \bibinfo{year}{2023}\natexlab{}.
\newblock \showarticletitle{Invigorating Ubuntu Ethics in AI for healthcare: Enabling equitable care}. In \bibinfo{booktitle}{\emph{Proceedings of the 2023 ACM Conference on Fairness, Accountability, and Transparency}}. \bibinfo{pages}{583--592}.
\newblock


\bibitem[Anderson and Bone(1995)]%
        {anderson1995first}
\bibfield{author}{\bibinfo{person}{Robert~B Anderson} {and} \bibinfo{person}{Robert~M Bone}.} \bibinfo{year}{1995}\natexlab{}.
\newblock \showarticletitle{First Nations economic development: a contingency perspective}.
\newblock \bibinfo{journal}{\emph{Canadian Geographer/Le G{\'e}ographe Canadien}} \bibinfo{volume}{39}, \bibinfo{number}{2} (\bibinfo{year}{1995}), \bibinfo{pages}{120--130}.
\newblock


\bibitem[Bender(2019)]%
        {bender2019benderrule}
\bibfield{author}{\bibinfo{person}{Emily Bender}.} \bibinfo{year}{2019}\natexlab{}.
\newblock \showarticletitle{The\# benderrule: On naming the languages we study and why it matters}.
\newblock \bibinfo{journal}{\emph{The Gradient}}  \bibinfo{volume}{14} (\bibinfo{year}{2019}).
\newblock


\bibitem[Bender et~al\mbox{.}(2021)]%
        {bender2021dangers}
\bibfield{author}{\bibinfo{person}{Emily~M Bender}, \bibinfo{person}{Timnit Gebru}, \bibinfo{person}{Angelina McMillan-Major}, {and} \bibinfo{person}{Shmargaret Shmitchell}.} \bibinfo{year}{2021}\natexlab{}.
\newblock \showarticletitle{On the dangers of stochastic parrots: Can language models be too big?}. In \bibinfo{booktitle}{\emph{Proceedings of the 2021 ACM conference on fairness, accountability, and transparency}}. \bibinfo{pages}{610--623}.
\newblock


\bibitem[Benjamin(2023)]%
        {benjamin2023race}
\bibfield{author}{\bibinfo{person}{Ruha Benjamin}.} \bibinfo{year}{2023}\natexlab{}.
\newblock \showarticletitle{Race after technology}.
\newblock In \bibinfo{booktitle}{\emph{Social Theory Re-Wired}}. \bibinfo{publisher}{Routledge}, \bibinfo{pages}{405--415}.
\newblock


\bibitem[Bird(2020)]%
        {bird2020decolonising}
\bibfield{author}{\bibinfo{person}{Steven Bird}.} \bibinfo{year}{2020}\natexlab{}.
\newblock \showarticletitle{Decolonising speech and language technology}. In \bibinfo{booktitle}{\emph{Proceedings of the 28th international conference on computational linguistics}}. \bibinfo{pages}{3504--3519}.
\newblock


\bibitem[Bird(2022)]%
        {bird2022local}
\bibfield{author}{\bibinfo{person}{Steven Bird}.} \bibinfo{year}{2022}\natexlab{}.
\newblock \showarticletitle{Local languages, third spaces, and other high-resource scenarios}. In \bibinfo{booktitle}{\emph{Proceedings of the 60th Annual Meeting of the Association for Computational Linguistics (Volume 1: Long Papers)}}. \bibinfo{pages}{7817--7829}.
\newblock


\bibitem[Birhane(2020)]%
        {birhane2020algorithmic}
\bibfield{author}{\bibinfo{person}{Abeba Birhane}.} \bibinfo{year}{2020}\natexlab{}.
\newblock \showarticletitle{Algorithmic colonization of Africa}.
\newblock \bibinfo{journal}{\emph{SCRIPTed}}  \bibinfo{volume}{17} (\bibinfo{year}{2020}), \bibinfo{pages}{389}.
\newblock


\bibitem[Birhane(2021)]%
        {BIRHANE2021100205}
\bibfield{author}{\bibinfo{person}{Abeba Birhane}.} \bibinfo{year}{2021}\natexlab{}.
\newblock \showarticletitle{Algorithmic injustice: a relational ethics approach}.
\newblock \bibinfo{journal}{\emph{Patterns}} \bibinfo{volume}{2}, \bibinfo{number}{2} (\bibinfo{year}{2021}), \bibinfo{pages}{100205}.
\newblock
\showISSN{2666-3899}
\urldef\tempurl%
\url{https://doi.org/10.1016/j.patter.2021.100205}
\showDOI{\tempurl}


\bibitem[Birhane et~al\mbox{.}(2022)]%
        {BirhanePower}
\bibfield{author}{\bibinfo{person}{Abeba Birhane}, \bibinfo{person}{William Isaac}, \bibinfo{person}{Vinodkumar Prabhakaran}, \bibinfo{person}{Mark Diaz}, \bibinfo{person}{Madeleine~Clare Elish}, \bibinfo{person}{Iason Gabriel}, {and} \bibinfo{person}{Shakir Mohamed}.} \bibinfo{year}{2022}\natexlab{}.
\newblock \showarticletitle{Power to the People? Opportunities and Challenges for Participatory AI}. In \bibinfo{booktitle}{\emph{Equity and Access in Algorithms, Mechanisms, and Optimization}} (Arlington, VA, USA) \emph{(\bibinfo{series}{EAAMO '22})}. \bibinfo{publisher}{Association for Computing Machinery}, \bibinfo{address}{New York, NY, USA}, Article \bibinfo{articleno}{6}, \bibinfo{numpages}{8}~pages.
\newblock
\showISBNx{9781450394772}
\urldef\tempurl%
\url{https://doi.org/10.1145/3551624.3555290}
\showDOI{\tempurl}


\bibitem[Blench(2012)]%
        {blench2012atlas}
\bibfield{author}{\bibinfo{person}{Roger Blench}.} \bibinfo{year}{2012}\natexlab{}.
\newblock \bibinfo{booktitle}{\emph{An atlas of Nigerian languages}}.
\newblock \bibinfo{publisher}{Kay Williamson Educational Foundation Oxford}.
\newblock


\bibitem[Blommaert(2004)]%
        {blommaert2004writing}
\bibfield{author}{\bibinfo{person}{Jan Blommaert}.} \bibinfo{year}{2004}\natexlab{}.
\newblock \showarticletitle{Writing as a problem: African grassroots writing, economies of literacy, and globalization}.
\newblock \bibinfo{journal}{\emph{Language in society}} \bibinfo{volume}{33}, \bibinfo{number}{5} (\bibinfo{year}{2004}), \bibinfo{pages}{643--671}.
\newblock


\bibitem[Bourdieu(1990)]%
        {bourdieu1990logic}
\bibfield{author}{\bibinfo{person}{Pierre Bourdieu}.} \bibinfo{year}{1990}\natexlab{}.
\newblock \bibinfo{booktitle}{\emph{The logic of practice}}.
\newblock \bibinfo{publisher}{Stanford university press}.
\newblock


\bibitem[Bright et~al\mbox{.}(2022)]%
        {bright2022stability}
\bibfield{author}{\bibinfo{person}{Liam~Kofi Bright}, \bibinfo{person}{Nathan Gabriel}, \bibinfo{person}{Cailin O'Connor}, {and} \bibinfo{person}{Olufemi Taiwo}.} \bibinfo{year}{2022}\natexlab{}.
\newblock \showarticletitle{On the stability of racial capitalism}.
\newblock  (\bibinfo{year}{2022}).
\newblock


\bibitem[Brown et~al\mbox{.}(2020)]%
        {brown2020language}
\bibfield{author}{\bibinfo{person}{Tom Brown}, \bibinfo{person}{Benjamin Mann}, \bibinfo{person}{Nick Ryder}, \bibinfo{person}{Melanie Subbiah}, \bibinfo{person}{Jared~D Kaplan}, \bibinfo{person}{Prafulla Dhariwal}, \bibinfo{person}{Arvind Neelakantan}, \bibinfo{person}{Pranav Shyam}, \bibinfo{person}{Girish Sastry}, \bibinfo{person}{Amanda Askell}, {et~al\mbox{.}}} \bibinfo{year}{2020}\natexlab{}.
\newblock \showarticletitle{Language models are few-shot learners}.
\newblock \bibinfo{journal}{\emph{Advances in neural information processing systems}}  \bibinfo{volume}{33} (\bibinfo{year}{2020}), \bibinfo{pages}{1877--1901}.
\newblock


\bibitem[Burke(2004)]%
        {burke2004languages}
\bibfield{author}{\bibinfo{person}{Peter Burke}.} \bibinfo{year}{2004}\natexlab{}.
\newblock \bibinfo{booktitle}{\emph{Languages and communities in early modern Europe}}.
\newblock \bibinfo{publisher}{Cambridge University Press}.
\newblock


\bibitem[Burkhart(2019)]%
        {burkhart2019indigenizing}
\bibfield{author}{\bibinfo{person}{Brian Burkhart}.} \bibinfo{year}{2019}\natexlab{}.
\newblock \bibinfo{booktitle}{\emph{Indigenizing philosophy through the land: A trickster methodology for decolonizing environmental ethics and Indigenous futures}}.
\newblock \bibinfo{publisher}{MSU Press}.
\newblock


\bibitem[Butler(1999)]%
        {butler1999coconstructing}
\bibfield{author}{\bibinfo{person}{Judith Butler}.} \bibinfo{year}{1999}\natexlab{}.
\newblock \bibinfo{booktitle}{\emph{Performativity’s social magic}}.
\newblock \bibinfo{publisher}{Oxford: Blackwell}. 113–128 pages.
\newblock


\bibitem[Caron et~al\mbox{.}(2023)]%
        {caron2023partnering}
\bibfield{author}{\bibinfo{person}{Nadine~R Caron}, \bibinfo{person}{Wilf Adam}, \bibinfo{person}{Kate Anderson}, \bibinfo{person}{Brooke~T Boswell}, \bibinfo{person}{Meck Chongo}, \bibinfo{person}{Viktor Deineko}, \bibinfo{person}{Alexanne Dick}, \bibinfo{person}{Shannon~E Hall}, \bibinfo{person}{Jessica~T Hatcher}, \bibinfo{person}{Patricia Howard}, {et~al\mbox{.}}} \bibinfo{year}{2023}\natexlab{}.
\newblock \showarticletitle{Partnering with First Nations in Northern British Columbia Canada to Reduce Inequity in Access to Genomic Research}.
\newblock \bibinfo{journal}{\emph{International Journal of Environmental Research and Public Health}} \bibinfo{volume}{20}, \bibinfo{number}{10} (\bibinfo{year}{2023}), \bibinfo{pages}{5783}.
\newblock


\bibitem[Charles et~al\mbox{.}(2020)]%
        {charles2020community}
\bibfield{author}{\bibinfo{person}{Anthony Charles}, \bibinfo{person}{Laura Loucks}, \bibinfo{person}{Fikret Berkes}, {and} \bibinfo{person}{Derek Armitage}.} \bibinfo{year}{2020}\natexlab{}.
\newblock \showarticletitle{Community science: A typology and its implications for governance of social-ecological systems}.
\newblock \bibinfo{journal}{\emph{Environmental Science \& Policy}}  \bibinfo{volume}{106} (\bibinfo{year}{2020}), \bibinfo{pages}{77--86}.
\newblock


\bibitem[Chiblow and Meighan(2022)]%
        {chiblow2022language}
\bibfield{author}{\bibinfo{person}{Susan Chiblow} {and} \bibinfo{person}{Paul~J Meighan}.} \bibinfo{year}{2022}\natexlab{}.
\newblock \showarticletitle{Language is land, land is language: The importance of Indigenous languages}.
\newblock \bibinfo{journal}{\emph{Human Geography}} \bibinfo{volume}{15}, \bibinfo{number}{2} (\bibinfo{year}{2022}), \bibinfo{pages}{206--210}.
\newblock


\bibitem[Chrupa{\l}a(2023)]%
        {chrupala2023putting}
\bibfield{author}{\bibinfo{person}{Grzegorz Chrupa{\l}a}.} \bibinfo{year}{2023}\natexlab{}.
\newblock \showarticletitle{Putting Natural in Natural Language Processing}.
\newblock \bibinfo{journal}{\emph{arXiv preprint arXiv:2305.04572}} (\bibinfo{year}{2023}).
\newblock


\bibitem[Collins and Bilge(2020)]%
        {collins2020intersectionality}
\bibfield{author}{\bibinfo{person}{Patricia~Hill Collins} {and} \bibinfo{person}{Sirma Bilge}.} \bibinfo{year}{2020}\natexlab{}.
\newblock \bibinfo{booktitle}{\emph{Intersectionality}}.
\newblock \bibinfo{publisher}{John Wiley \& Sons}.
\newblock


\bibitem[Cooper et~al\mbox{.}(2024)]%
        {cooper2024}
\bibfield{author}{\bibinfo{person}{Ned Cooper}, \bibinfo{person}{Courtney Heldreth}, {and} \bibinfo{person}{Ben Hutchinson}.} \bibinfo{year}{2024}\natexlab{}.
\newblock \showarticletitle{``It's \emph{how} you do things that matters'': Attending to Process to Better Serve Indigenous Communities with Language Technologies}.
\newblock \bibinfo{journal}{\emph{Proceedings of The 18th Conference of the European Chapter of the Association for Computational Linguistics ({EACL} 2024)}} (\bibinfo{year}{2024}).
\newblock


\bibitem[CUERVO(2002)]%
        {cuervo2002instituto}
\bibfield{author}{\bibinfo{person}{OBRAS IN{\'E}DITAS DE RUFINO~J CUERVO}.} \bibinfo{year}{2002}\natexlab{}.
\newblock \showarticletitle{Instituto Caro y Cuervo}.
\newblock \bibinfo{journal}{\emph{Diccionario de construcci{\'o}n y r{\'e}gimen de la lengua castellana}} (\bibinfo{year}{2002}).
\newblock


\bibitem[Dalby(2003)]%
        {dalby2003language}
\bibfield{author}{\bibinfo{person}{Andrew Dalby}.} \bibinfo{year}{2003}\natexlab{}.
\newblock \bibinfo{booktitle}{\emph{Language in danger: The loss of linguistic diversity and the threat to our future}}.
\newblock \bibinfo{publisher}{Columbia University Press}.
\newblock


\bibitem[Dance(2010)]%
        {39248288bedc4e18af990501e6b9e285}
\bibfield{author}{\bibinfo{person}{Lory Dance}.} \bibinfo{year}{2010}\natexlab{}.
\newblock \bibinfo{title}{Struggles of the Disenfranchised: Commonalities Among Native Americans, Black Americans, and Palestinians}.
\newblock
\newblock


\bibitem[Davis et~al\mbox{.}(2021)]%
        {davis2021algorithmic}
\bibfield{author}{\bibinfo{person}{Jenny~L Davis}, \bibinfo{person}{Apryl Williams}, {and} \bibinfo{person}{Michael~W Yang}.} \bibinfo{year}{2021}\natexlab{}.
\newblock \showarticletitle{Algorithmic reparation}.
\newblock \bibinfo{journal}{\emph{Big Data \& Society}} \bibinfo{volume}{8}, \bibinfo{number}{2} (\bibinfo{year}{2021}), \bibinfo{pages}{20539517211044808}.
\newblock


\bibitem[Deloria(2001)]%
        {deloria2001american}
\bibfield{author}{\bibinfo{person}{Vine Deloria}.} \bibinfo{year}{2001}\natexlab{}.
\newblock \showarticletitle{American Indian metaphysics}.
\newblock \bibinfo{journal}{\emph{Winds of Change}} (\bibinfo{year}{2001}), \bibinfo{pages}{49--67}.
\newblock


\bibitem[Denton et~al\mbox{.}(2021)]%
        {denton2021genealogy}
\bibfield{author}{\bibinfo{person}{Emily Denton}, \bibinfo{person}{Alex Hanna}, \bibinfo{person}{Razvan Amironesei}, \bibinfo{person}{Andrew Smart}, {and} \bibinfo{person}{Hilary Nicole}.} \bibinfo{year}{2021}\natexlab{}.
\newblock \showarticletitle{On the genealogy of machine learning datasets: A critical history of ImageNet}.
\newblock \bibinfo{journal}{\emph{Big Data \& Society}} \bibinfo{volume}{8}, \bibinfo{number}{2} (\bibinfo{year}{2021}), \bibinfo{pages}{20539517211035955}.
\newblock


\bibitem[Dev et~al\mbox{.}(2023)]%
        {dev2023building}
\bibfield{author}{\bibinfo{person}{Sunipa Dev}, \bibinfo{person}{Jaya Goyal}, \bibinfo{person}{Dinesh Tewari}, \bibinfo{person}{Shachi Dave}, {and} \bibinfo{person}{Vinodkumar Prabhakaran}.} \bibinfo{year}{2023}\natexlab{}.
\newblock \showarticletitle{Building Socio-culturally Inclusive Stereotype Resources with Community Engagement}.
\newblock \bibinfo{journal}{\emph{arXiv preprint arXiv:2307.10514}} (\bibinfo{year}{2023}).
\newblock


\bibitem[Dobrin et~al\mbox{.}(2014)]%
        {dobrin2014dying}
\bibfield{author}{\bibinfo{person}{Lise Dobrin}, \bibinfo{person}{Peter~K Austin}, {and} \bibinfo{person}{David Nathan}.} \bibinfo{year}{2014}\natexlab{}.
\newblock \showarticletitle{Dying to be counted: The commodification of endangered languages in documentary linguistics}.
\newblock \bibinfo{journal}{\emph{Language documentation and description}}  \bibinfo{volume}{6} (\bibinfo{year}{2014}).
\newblock


\bibitem[Dobrin and Good(2009)]%
        {dobrin2009practical}
\bibfield{author}{\bibinfo{person}{Lise~M Dobrin} {and} \bibinfo{person}{Jeff Good}.} \bibinfo{year}{2009}\natexlab{}.
\newblock \showarticletitle{Practical language development: Whose mission?}
\newblock \bibinfo{journal}{\emph{Language}} \bibinfo{volume}{85}, \bibinfo{number}{3} (\bibinfo{year}{2009}), \bibinfo{pages}{619--629}.
\newblock


\bibitem[Ducel et~al\mbox{.}(2022)]%
        {ducel2022we}
\bibfield{author}{\bibinfo{person}{Fanny Ducel}, \bibinfo{person}{Kar{\"e}n Fort}, \bibinfo{person}{Ga{\"e}l Lejeune}, {and} \bibinfo{person}{Yves Lepage}.} \bibinfo{year}{2022}\natexlab{}.
\newblock \showarticletitle{Do we Name the Languages we Study? The\# BenderRule in LREC and ACL articles}. In \bibinfo{booktitle}{\emph{Proceedings of the Thirteenth Language Resources and Evaluation Conference}}. \bibinfo{pages}{564--573}.
\newblock


\bibitem[Eberhard et~al\mbox{.}(2019)]%
        {eberhard2019summary}
\bibfield{author}{\bibinfo{person}{David~M Eberhard}, \bibinfo{person}{Gary~F Simons}, {and} \bibinfo{person}{Charles~D Fennig}.} \bibinfo{year}{2019}\natexlab{}.
\newblock \showarticletitle{Summary by language size}.
\newblock \bibinfo{journal}{\emph{SIL International, Ethnologue}} (\bibinfo{year}{2019}).
\newblock


\bibitem[Emezue and Dossou(2020)]%
        {emezue2020lanfrica}
\bibfield{author}{\bibinfo{person}{Chris~C Emezue} {and} \bibinfo{person}{Bonaventure~FP Dossou}.} \bibinfo{year}{2020}\natexlab{}.
\newblock \showarticletitle{Lanfrica: A participatory approach to documenting machine translation research on African languages}.
\newblock \bibinfo{journal}{\emph{arXiv preprint arXiv:2008.07302}} (\bibinfo{year}{2020}).
\newblock


\bibitem[Ensmenger(2021)]%
        {ensmenger2021cloud}
\bibfield{author}{\bibinfo{person}{Nathan Ensmenger}.} \bibinfo{year}{2021}\natexlab{}.
\newblock \bibinfo{booktitle}{\emph{"The Cloud is a Factory," in Thomas Mullaney, Benjamin Peters, Mar Hicks, and Kavita Philip, eds. Your Computer is On Fire}}.
\newblock \bibinfo{publisher}{MIT Press}, Chapter~1, \bibinfo{pages}{43--45}.
\newblock


\bibitem[Eriksen(1992)]%
        {eriksen1992linguistic}
\bibfield{author}{\bibinfo{person}{Thomas~Hylland Eriksen}.} \bibinfo{year}{1992}\natexlab{}.
\newblock \showarticletitle{Linguistic hegemony and minority resistance}.
\newblock \bibinfo{journal}{\emph{Journal of Peace Research}} \bibinfo{volume}{29}, \bibinfo{number}{3} (\bibinfo{year}{1992}), \bibinfo{pages}{313--332}.
\newblock


\bibitem[Faloyin(2022)]%
        {Faloyin2022}
\bibfield{author}{\bibinfo{person}{Dipo Faloyin}.} \bibinfo{year}{2022}\natexlab{}.
\newblock \bibinfo{booktitle}{\emph{Africa Is Not a Country: Notes on a Bright Continent}}.
\newblock \bibinfo{publisher}{W. W. Norton \& Company}.
\newblock
\showISBNx{978-0-393-88153-0}


\bibitem[Gade(2012)]%
        {Gade2012}
\bibfield{author}{\bibinfo{person}{Christian~B.N. Gade}.} \bibinfo{year}{2012}\natexlab{}.
\newblock \showarticletitle{What is Ubuntu? Different Interpretations among South Africans of African Descent}.
\newblock \bibinfo{journal}{\emph{South African Journal of Philosophy}}  \bibinfo{volume}{31} (\bibinfo{date}{1} \bibinfo{year}{2012}), \bibinfo{pages}{484--503}.
\newblock
Issue 3.
\showISSN{0258-0136}
\urldef\tempurl%
\url{https://doi.org/10.1080/02580136.2012.10751789}
\showDOI{\tempurl}


\bibitem[Gal({[n.\,d.]})]%
        {gal2006migration}
\bibfield{author}{\bibinfo{person}{Susan Gal}.} \bibinfo{year}{[n.\,d.]}\natexlab{}.
\newblock \showarticletitle{Migration, minorities and multilingualism: Language ideologies in Europe}.
\newblock In \bibinfo{booktitle}{\emph{Language ideologies, policies and practices: Language and the future of Europe}}. \bibinfo{publisher}{Springer}, \bibinfo{pages}{13--27}.
\newblock


\bibitem[Geertz(1973)]%
        {geertz1973thick}
\bibfield{author}{\bibinfo{person}{Clifford Geertz}.} \bibinfo{year}{1973}\natexlab{}.
\newblock \bibinfo{booktitle}{\emph{Thick description}}.
\newblock


\bibitem[Gilbert(2007)]%
        {gilbert2007indigenous}
\bibfield{author}{\bibinfo{person}{J{\'e}r{\'e}mie Gilbert}.} \bibinfo{year}{2007}\natexlab{}.
\newblock \showarticletitle{Indigenous rights in the making: The United Nations declaration on the rights of indigenous peoples}.
\newblock \bibinfo{journal}{\emph{Int'l J. on Minority \& Group Rts.}}  \bibinfo{volume}{14} (\bibinfo{year}{2007}), \bibinfo{pages}{207}.
\newblock


\bibitem[Gitelman(2013)]%
        {gitelman2013raw}
\bibfield{author}{\bibinfo{person}{Lisa Gitelman}.} \bibinfo{year}{2013}\natexlab{}.
\newblock \bibinfo{booktitle}{\emph{Raw data is an oxymoron}}.
\newblock \bibinfo{publisher}{MIT press}.
\newblock


\bibitem[Hagerty(2023)]%
        {hagerty2023climate_divide}
\bibfield{author}{\bibinfo{person}{Colleen Hagerty}.} \bibinfo{year}{July/August 2023}\natexlab{}.
\newblock \showarticletitle{How climate vulnerability and the digital divide are linked}.
\newblock \bibinfo{journal}{\emph{MIT Technology Review}} (\bibinfo{year}{July/August 2023}).
\newblock


\bibitem[Hoffmann(2021)]%
        {hoffmann2021terms}
\bibfield{author}{\bibinfo{person}{Anna~Lauren Hoffmann}.} \bibinfo{year}{2021}\natexlab{}.
\newblock \showarticletitle{Terms of inclusion: Data, discourse, violence}.
\newblock \bibinfo{journal}{\emph{New Media \& Society}} \bibinfo{volume}{23}, \bibinfo{number}{12} (\bibinfo{year}{2021}), \bibinfo{pages}{3539--3556}.
\newblock


\bibitem[Holton et~al\mbox{.}(2022)]%
        {holton2022indigenous}
\bibfield{author}{\bibinfo{person}{Gary Holton}, \bibinfo{person}{Wesley~Y Leonard}, {and} \bibinfo{person}{Peter~L Pulsifer}.} \bibinfo{year}{2022}\natexlab{}.
\newblock \showarticletitle{Indigenous peoples, ethics, and linguistic data}.
\newblock \bibinfo{journal}{\emph{BEREZ-KROEKER Andrea L.; MCDONNELL, Bradley; KOLLER, Eve}} (\bibinfo{year}{2022}), \bibinfo{pages}{49--60}.
\newblock


\bibitem[Jauhiainen et~al\mbox{.}(2019)]%
        {jauhiainen2019automatic}
\bibfield{author}{\bibinfo{person}{Tommi Jauhiainen}, \bibinfo{person}{Marco Lui}, \bibinfo{person}{Marcos Zampieri}, \bibinfo{person}{Timothy Baldwin}, {and} \bibinfo{person}{Krister Lind{\'e}n}.} \bibinfo{year}{2019}\natexlab{}.
\newblock \showarticletitle{Automatic language identification in texts: A survey}.
\newblock \bibinfo{journal}{\emph{Journal of Artificial Intelligence Research}}  \bibinfo{volume}{65} (\bibinfo{year}{2019}), \bibinfo{pages}{675--782}.
\newblock


\bibitem[Johnson et~al\mbox{.}(2022)]%
        {johnson2022ghost}
\bibfield{author}{\bibinfo{person}{Rebecca~L Johnson}, \bibinfo{person}{Giada Pistilli}, \bibinfo{person}{Natalia Men{\'e}dez-Gonz{\'a}lez}, \bibinfo{person}{Leslye Denisse~Dias Duran}, \bibinfo{person}{Enrico Panai}, \bibinfo{person}{Julija Kalpokiene}, {and} \bibinfo{person}{Donald~Jay Bertulfo}.} \bibinfo{year}{2022}\natexlab{}.
\newblock \showarticletitle{The Ghost in the Machine has an American accent: value conflict in GPT-3}.
\newblock \bibinfo{journal}{\emph{arXiv preprint arXiv:2203.07785}} (\bibinfo{year}{2022}).
\newblock


\bibitem[Kapania et~al\mbox{.}(2023)]%
        {kapania2023hunt}
\bibfield{author}{\bibinfo{person}{Shivani Kapania}, \bibinfo{person}{Alex~S Taylor}, {and} \bibinfo{person}{Ding Wang}.} \bibinfo{year}{2023}\natexlab{}.
\newblock \showarticletitle{A hunt for the Snark: Annotator Diversity in Data Practices}. In \bibinfo{booktitle}{\emph{Proceedings of the 2023 CHI Conference on Human Factors in Computing Systems}}. \bibinfo{pages}{1--15}.
\newblock


\bibitem[Keegan(2017)]%
        {keegan2017maori}
\bibfield{author}{\bibinfo{person}{Peter~J Keegan}.} \bibinfo{year}{2017}\natexlab{}.
\newblock \showarticletitle{M{\"a}ori Dialect Issues and M{\"a}ori Language Ideologies in the Revitalisation Era}.
\newblock \bibinfo{journal}{\emph{MAI Journal}} \bibinfo{volume}{6}, \bibinfo{number}{2} (\bibinfo{year}{2017}), \bibinfo{pages}{129--142}.
\newblock


\bibitem[Kirk et~al\mbox{.}(2023)]%
        {kirk2023understanding}
\bibfield{author}{\bibinfo{person}{Robert Kirk}, \bibinfo{person}{Ishita Mediratta}, \bibinfo{person}{Christoforos Nalmpantis}, \bibinfo{person}{Jelena Luketina}, \bibinfo{person}{Eric Hambro}, \bibinfo{person}{Edward Grefenstette}, {and} \bibinfo{person}{Roberta Raileanu}.} \bibinfo{year}{2023}\natexlab{}.
\newblock \showarticletitle{Understanding the Effects of RLHF on LLM Generalisation and Diversity}.
\newblock \bibinfo{journal}{\emph{arXiv preprint arXiv:2310.06452}} (\bibinfo{year}{2023}).
\newblock


\bibitem[Kotut and McCrickard(2022)]%
        {kotutmck}
\bibfield{author}{\bibinfo{person}{Lindah Kotut} {and} \bibinfo{person}{D~Scott McCrickard}.} \bibinfo{year}{2022}\natexlab{}.
\newblock \showarticletitle{Winds of Change: Seeking, Preserving, and Retelling Indigenous Knowledge Through Self-Organized Online Communities}.
\newblock  (\bibinfo{year}{2022}), \bibinfo{pages}{1--15}.
\newblock


\bibitem[Kr{\"a}mer et~al\mbox{.}(2022)]%
        {kramer2022language}
\bibfield{author}{\bibinfo{person}{Philipp Kr{\"a}mer}, \bibinfo{person}{Ulrike Vogl}, {and} \bibinfo{person}{Leena Kolehmainen}.} \bibinfo{year}{2022}\natexlab{}.
\newblock \showarticletitle{What is “Language Making”?}
\newblock \bibinfo{journal}{\emph{International Journal of the Sociology of Language}} \bibinfo{volume}{2022}, \bibinfo{number}{274} (\bibinfo{year}{2022}), \bibinfo{pages}{1--27}.
\newblock


\bibitem[Kuhn et~al\mbox{.}(2020)]%
        {kuhn2020indigenous}
\bibfield{author}{\bibinfo{person}{Roland Kuhn}, \bibinfo{person}{Fineen Davis}, \bibinfo{person}{Alain D{\'e}silets}, \bibinfo{person}{Eric Joanis}, \bibinfo{person}{Anna Kazantseva}, \bibinfo{person}{Rebecca Knowles}, \bibinfo{person}{Patrick Littell}, \bibinfo{person}{Delaney Lothian}, \bibinfo{person}{Aidan Pine}, \bibinfo{person}{Caroline~Running Wolf}, {et~al\mbox{.}}} \bibinfo{year}{2020}\natexlab{}.
\newblock \showarticletitle{The Indigenous Languages Technology project at NRC Canada: An empowerment-oriented approach to developing language software}. In \bibinfo{booktitle}{\emph{Proceedings of the 28th international conference on computational linguistics}}. \bibinfo{pages}{5866--5878}.
\newblock


\bibitem[Kukutai et~al\mbox{.}(2020)]%
        {kukutai2020indigenous}
\bibfield{author}{\bibinfo{person}{Tahu Kukutai}, \bibinfo{person}{Stephanie~Russo Carroll}, {and} \bibinfo{person}{Maggie Walter}.} \bibinfo{year}{2020}\natexlab{}.
\newblock \showarticletitle{Indigenous data sovereignty}.
\newblock  (\bibinfo{year}{2020}).
\newblock


\bibitem[Landauer(1999)]%
        {landauer1999latent}
\bibfield{author}{\bibinfo{person}{Thomas~K Landauer}.} \bibinfo{year}{1999}\natexlab{}.
\newblock \showarticletitle{Latent semantic analysis: A theory of the psychology of language and mind}.
\newblock  (\bibinfo{year}{1999}).
\newblock


\bibitem[Leonard(2017)]%
        {leonard2017producing}
\bibfield{author}{\bibinfo{person}{Wesley~Y Leonard}.} \bibinfo{year}{2017}\natexlab{}.
\newblock \showarticletitle{Producing language reclamation by decolonising ‘language’}.
\newblock \bibinfo{journal}{\emph{Language documentation and description}}  \bibinfo{volume}{14} (\bibinfo{year}{2017}).
\newblock


\bibitem[Liu et~al\mbox{.}(2023)]%
        {liu2023summary}
\bibfield{author}{\bibinfo{person}{Yiheng Liu}, \bibinfo{person}{Tianle Han}, \bibinfo{person}{Siyuan Ma}, \bibinfo{person}{Jiayue Zhang}, \bibinfo{person}{Yuanyuan Yang}, \bibinfo{person}{Jiaming Tian}, \bibinfo{person}{Hao He}, \bibinfo{person}{Antong Li}, \bibinfo{person}{Mengshen He}, \bibinfo{person}{Zhengliang Liu}, {et~al\mbox{.}}} \bibinfo{year}{2023}\natexlab{}.
\newblock \showarticletitle{Summary of chatgpt/gpt-4 research and perspective towards the future of large language models}.
\newblock \bibinfo{journal}{\emph{arXiv preprint arXiv:2304.01852}} (\bibinfo{year}{2023}).
\newblock


\bibitem[L{\"u}pke(2011)]%
        {lupke2011orthography}
\bibfield{author}{\bibinfo{person}{Friederike L{\"u}pke}.} \bibinfo{year}{2011}\natexlab{}.
\newblock \bibinfo{booktitle}{\emph{Orthography development}}.
\newblock \bibinfo{publisher}{Cambridge University Press}.
\newblock


\bibitem[L{\"u}pke(2015)]%
        {lupke2015ideologies}
\bibfield{author}{\bibinfo{person}{Friederike L{\"u}pke}.} \bibinfo{year}{2015}\natexlab{}.
\newblock \showarticletitle{Ideologies and typologies of language endangerment in Africa}.
\newblock \bibinfo{journal}{\emph{Language documentation and endangerment in Africa}}  \bibinfo{volume}{16} (\bibinfo{year}{2015}), \bibinfo{pages}{59--105}.
\newblock


\bibitem[Maffi(2002)]%
        {maffi2002endangered}
\bibfield{author}{\bibinfo{person}{Luisa Maffi}.} \bibinfo{year}{2002}\natexlab{}.
\newblock \showarticletitle{Endangered languages, endangered knowledge}.
\newblock \bibinfo{journal}{\emph{International Social Science Journal}} \bibinfo{volume}{54}, \bibinfo{number}{173} (\bibinfo{year}{2002}), \bibinfo{pages}{385--393}.
\newblock


\bibitem[Mager et~al\mbox{.}(2023)]%
        {mager2023ethical}
\bibfield{author}{\bibinfo{person}{Manuel Mager}, \bibinfo{person}{Elisabeth Mager}, \bibinfo{person}{Katharina Kann}, {and} \bibinfo{person}{Ngoc~Thang Vu}.} \bibinfo{year}{2023}\natexlab{}.
\newblock \showarticletitle{Ethical Considerations for Machine Translation of Indigenous Languages: Giving a Voice to the Speakers}.
\newblock \bibinfo{journal}{\emph{arXiv preprint arXiv:2305.19474}} (\bibinfo{year}{2023}).
\newblock


\bibitem[Marivate et~al\mbox{.}(2020)]%
        {marivate2020investigating}
\bibfield{author}{\bibinfo{person}{Vukosi Marivate}, \bibinfo{person}{Tshephisho Sefara}, \bibinfo{person}{Vongani Chabalala}, \bibinfo{person}{Keamogetswe Makhaya}, \bibinfo{person}{Tumisho Mokgonyane}, \bibinfo{person}{Rethabile Mokoena}, {and} \bibinfo{person}{Abiodun Modupe}.} \bibinfo{year}{2020}\natexlab{}.
\newblock \showarticletitle{Investigating an approach for low resource language dataset creation, curation and classification: Setswana and Sepedi}.
\newblock \bibinfo{journal}{\emph{arXiv preprint arXiv:2003.04986}} (\bibinfo{year}{2020}).
\newblock


\bibitem[McGregor et~al\mbox{.}(2019)]%
        {mcgregor2019international}
\bibfield{author}{\bibinfo{person}{Lorna McGregor}, \bibinfo{person}{Daragh Murray}, {and} \bibinfo{person}{Vivian Ng}.} \bibinfo{year}{2019}\natexlab{}.
\newblock \showarticletitle{International human rights law as a framework for algorithmic accountability}.
\newblock \bibinfo{journal}{\emph{International \& Comparative Law Quarterly}} \bibinfo{volume}{68}, \bibinfo{number}{2} (\bibinfo{year}{2019}), \bibinfo{pages}{309--343}.
\newblock


\bibitem[McKercher(2020)]%
        {mckercher2020beyond}
\bibfield{author}{\bibinfo{person}{Kelly~Ann McKercher}.} \bibinfo{year}{2020}\natexlab{}.
\newblock \showarticletitle{Beyond sticky notes}.
\newblock \bibinfo{journal}{\emph{Doing co-design for Real: Mindsets, Methods, and Movements, 1st Edn. Sydney, NSW: Beyond Sticky Notes}} (\bibinfo{year}{2020}).
\newblock


\bibitem[McWhorter(2014)]%
        {mcwhorter2014hoax}
\bibfield{author}{\bibinfo{person}{John McWhorter}.} \bibinfo{year}{2014}\natexlab{}.
\newblock \bibinfo{booktitle}{\emph{The Language Hoax: Why The World Looks The Same In Any Language}}.
\newblock \bibinfo{publisher}{Oxford University Press}.
\newblock


\bibitem[Medina(2012)]%
        {medina2012epistemology}
\bibfield{author}{\bibinfo{person}{Jos{\'e} Medina}.} \bibinfo{year}{2012}\natexlab{}.
\newblock \bibinfo{booktitle}{\emph{The epistemology of resistance: Gender and racial oppression, epistemic injustice, and resistant imaginations}}.
\newblock \bibinfo{publisher}{Oxford University Press}.
\newblock


\bibitem[Meighan(2021)]%
        {meighan2021decolonizing}
\bibfield{author}{\bibinfo{person}{Paul~J Meighan}.} \bibinfo{year}{2021}\natexlab{}.
\newblock \showarticletitle{Decolonizing English: A proposal for implementing alternative ways of knowing and being in education}.
\newblock \bibinfo{journal}{\emph{Diaspora, Indigenous, and Minority Education}} \bibinfo{volume}{15}, \bibinfo{number}{2} (\bibinfo{year}{2021}), \bibinfo{pages}{77--83}.
\newblock


\bibitem[Mihas et~al\mbox{.}(2013)]%
        {mihas2013responses}
\bibfield{author}{\bibinfo{person}{Elena Mihas}, \bibinfo{person}{Bernard Perley}, \bibinfo{person}{Gabriel Rei-Doval}, {and} \bibinfo{person}{Kathleen Wheatley}.} \bibinfo{year}{2013}\natexlab{}.
\newblock \bibinfo{booktitle}{\emph{Responses to Language Endangerment: In honor of Mickey Noonan. New directions in language documentation and language revitalization}}. Vol.~\bibinfo{volume}{142}.
\newblock \bibinfo{publisher}{John Benjamins Publishing}.
\newblock


\bibitem[Milliere and Buckner(2024)]%
        {Milliere2024API}
\bibfield{author}{\bibinfo{person}{Raphael Milliere} {and} \bibinfo{person}{Cameron Buckner}.} \bibinfo{year}{2024}\natexlab{}.
\newblock \showarticletitle{A Philosophical Introduction to Language Models -- Part I: Continuity With Classic Debates}.
\newblock
\urldef\tempurl%
\url{https://api.semanticscholar.org/CorpusID:266844364}
\showURL{%
\tempurl}


\bibitem[Moiloa(2023)]%
        {Moiloa_2023}
\bibfield{author}{\bibinfo{person}{Pelonomi Moiloa}.} \bibinfo{year}{2023}\natexlab{}.
\newblock \bibinfo{title}{How ai can keep disappearing languages alive}.
\newblock
\newblock
\urldef\tempurl%
\url{https://www.ted.com/talks/pelonomi_moiloa_how_ai_can_keep_disappearing_languages_alive}
\showURL{%
\tempurl}


\bibitem[Morey et~al\mbox{.}(2013)]%
        {morey2013language}
\bibfield{author}{\bibinfo{person}{Stephen Morey}, \bibinfo{person}{Mark~W Post}, {and} \bibinfo{person}{Victor~A Friedman}.} \bibinfo{year}{2013}\natexlab{}.
\newblock \showarticletitle{The language codes of ISO 639: A premature, ultimately unobtainable, and possibly damaging standardization}.
\newblock  (\bibinfo{year}{2013}).
\newblock


\bibitem[Network(2016)]%
        {network2016te}
\bibfield{author}{\bibinfo{person}{M{\=a}ori Data~Sovereignty Network}.} \bibinfo{year}{2016}\natexlab{}.
\newblock \bibinfo{title}{Te mana raraunga—M{\=a}ori Data Sovereignty Network charter}.
\newblock
\newblock


\bibitem[Noonan(2006)]%
        {noonan2006rise}
\bibfield{author}{\bibinfo{person}{Michael Noonan}.} \bibinfo{year}{2006}\natexlab{}.
\newblock \showarticletitle{The rise of ethnic consciousness and the politicization of language in West-Central Nepal}.
\newblock \bibinfo{journal}{\emph{Lesser-known languages of South Asia: Status and policies, case studies and applications of information technology}} (\bibinfo{year}{2006}), \bibinfo{pages}{161--174}.
\newblock


\bibitem[Orife et~al\mbox{.}(2020)]%
        {orife2020masakhane}
\bibfield{author}{\bibinfo{person}{Iroro Orife}, \bibinfo{person}{Julia Kreutzer}, \bibinfo{person}{Blessing Sibanda}, \bibinfo{person}{Daniel Whitenack}, \bibinfo{person}{Kathleen Siminyu}, \bibinfo{person}{Laura Martinus}, \bibinfo{person}{Jamiil~Toure Ali}, \bibinfo{person}{Jade Abbott}, \bibinfo{person}{Vukosi Marivate}, \bibinfo{person}{Salomon Kabongo}, {et~al\mbox{.}}} \bibinfo{year}{2020}\natexlab{}.
\newblock \showarticletitle{Masakhane--Machine Translation For Africa}.
\newblock \bibinfo{journal}{\emph{arXiv preprint arXiv:2003.11529}} (\bibinfo{year}{2020}).
\newblock


\bibitem[Parkins(2017)]%
        {parkins2017world}
\bibfield{author}{\bibinfo{person}{David Parkins}.} \bibinfo{year}{2017}\natexlab{}.
\newblock \showarticletitle{The world’s most valuable resource is no longer oil, but data}.
\newblock \bibinfo{journal}{\emph{The economist}}  \bibinfo{volume}{6} (\bibinfo{year}{2017}).
\newblock


\bibitem[Patrick(2004)]%
        {patrick2004speech}
\bibfield{author}{\bibinfo{person}{Peter~L Patrick}.} \bibinfo{year}{2004}\natexlab{}.
\newblock \showarticletitle{The speech community}.
\newblock \bibinfo{journal}{\emph{The handbook of language variation and change}} (\bibinfo{year}{2004}), \bibinfo{pages}{573--597}.
\newblock


\bibitem[Paveau(2006)]%
        {paveau2006prediscourse}
\bibfield{author}{\bibinfo{person}{Marie-Anne Paveau}.} \bibinfo{year}{2006}\natexlab{}.
\newblock \bibinfo{booktitle}{\emph{Prédiscours: Sens, Mémoire, cognition}}.
\newblock \bibinfo{publisher}{Presse Sorbonne Nouvelle}.
\newblock


\bibitem[Poibeau(2017)]%
        {poibeau2017machine}
\bibfield{author}{\bibinfo{person}{Thierry Poibeau}.} \bibinfo{year}{2017}\natexlab{}.
\newblock \bibinfo{booktitle}{\emph{Machine translation}}.
\newblock \bibinfo{publisher}{MIT Press}.
\newblock


\bibitem[Prabhakaran et~al\mbox{.}(2022)]%
        {prabhakaran2022human}
\bibfield{author}{\bibinfo{person}{Vinodkumar Prabhakaran}, \bibinfo{person}{Margaret Mitchell}, \bibinfo{person}{Timnit Gebru}, {and} \bibinfo{person}{Iason Gabriel}.} \bibinfo{year}{2022}\natexlab{}.
\newblock \showarticletitle{A human rights-based approach to responsible ai}.
\newblock \bibinfo{journal}{\emph{arXiv preprint arXiv:2210.02667}} (\bibinfo{year}{2022}).
\newblock


\bibitem[Pratap et~al\mbox{.}(2023)]%
        {pratap2023scaling}
\bibfield{author}{\bibinfo{person}{Vineel Pratap}, \bibinfo{person}{Andros Tjandra}, \bibinfo{person}{Bowen Shi}, \bibinfo{person}{Paden Tomasello}, \bibinfo{person}{Arun Babu}, \bibinfo{person}{Sayani Kundu}, \bibinfo{person}{Ali Elkahky}, \bibinfo{person}{Zhaoheng Ni}, \bibinfo{person}{Apoorv Vyas}, \bibinfo{person}{Maryam Fazel-Zarandi}, {et~al\mbox{.}}} \bibinfo{year}{2023}\natexlab{}.
\newblock \showarticletitle{Scaling speech technology to 1,000+ languages}.
\newblock \bibinfo{journal}{\emph{arXiv preprint arXiv:2305.13516}} (\bibinfo{year}{2023}).
\newblock


\bibitem[Rao and McMahan(2019)]%
        {rao2019natural}
\bibfield{author}{\bibinfo{person}{Delip Rao} {and} \bibinfo{person}{Brian McMahan}.} \bibinfo{year}{2019}\natexlab{}.
\newblock \bibinfo{booktitle}{\emph{Natural language processing with PyTorch: build intelligent language applications using deep learning}}.
\newblock \bibinfo{publisher}{" O'Reilly Media, Inc."}.
\newblock


\bibitem[Ruder(2022)]%
        {ruder2022statemultilingualai}
\bibfield{author}{\bibinfo{person}{Sebastian Ruder}.} \bibinfo{year}{2022}\natexlab{}.
\newblock \bibinfo{title}{{The State of Multilingual AI}}.
\newblock \bibinfo{howpublished}{\url{http://ruder.io/state-of-multilingual-ai/}}.
\newblock


\bibitem[Sambasivan et~al\mbox{.}(2021)]%
        {sambasivan2021everyone}
\bibfield{author}{\bibinfo{person}{Nithya Sambasivan}, \bibinfo{person}{Shivani Kapania}, \bibinfo{person}{Hannah Highfill}, \bibinfo{person}{Diana Akrong}, \bibinfo{person}{Praveen Paritosh}, {and} \bibinfo{person}{Lora~M Aroyo}.} \bibinfo{year}{2021}\natexlab{}.
\newblock \showarticletitle{“Everyone wants to do the model work, not the data work”: Data Cascades in High-Stakes AI}. In \bibinfo{booktitle}{\emph{proceedings of the 2021 CHI Conference on Human Factors in Computing Systems}}. \bibinfo{pages}{1--15}.
\newblock


\bibitem[Savci(2015)]%
        {savci2015culture}
\bibfield{author}{\bibinfo{person}{Evren Savci}.} \bibinfo{year}{2015}\natexlab{}.
\newblock \showarticletitle{Language and Social Knowledge}.
\newblock \bibinfo{journal}{\emph{Ethnography}} (\bibinfo{year}{2015}), \bibinfo{pages}{1–11}.
\newblock


\bibitem[Scham(2001)]%
        {scham2001archaeology}
\bibfield{author}{\bibinfo{person}{Sandra~Arnold Scham}.} \bibinfo{year}{2001}\natexlab{}.
\newblock \showarticletitle{The archaeology of the disenfranchised}.
\newblock \bibinfo{journal}{\emph{Journal of Archaeological Method and Theory}}  \bibinfo{volume}{8} (\bibinfo{year}{2001}), \bibinfo{pages}{183--213}.
\newblock


\bibitem[Schwartz(2022)]%
        {schwartz2022primum}
\bibfield{author}{\bibinfo{person}{Lane Schwartz}.} \bibinfo{year}{2022}\natexlab{}.
\newblock \showarticletitle{Primum Non Nocere: Before working with Indigenous data, the ACL must confront ongoing colonialism}. In \bibinfo{booktitle}{\emph{Proceedings of the 60th Annual Meeting of the Association for Computational Linguistics}}, Vol.~\bibinfo{volume}{2}.
\newblock


\bibitem[Skutnabb-Kangas and Phillipson(2006)]%
        {skutnabb2006linguistic}
\bibfield{author}{\bibinfo{person}{Tove Skutnabb-Kangas} {and} \bibinfo{person}{Robert Phillipson}.} \bibinfo{year}{2006}\natexlab{}.
\newblock \showarticletitle{Linguistic Rights}.
\newblock \bibinfo{journal}{\emph{Encyclopedia of Language and Linguistics}} (\bibinfo{year}{2006}), \bibinfo{pages}{212--215}.
\newblock


\bibitem[Smith(2021)]%
        {smith2021decolonizing}
\bibfield{author}{\bibinfo{person}{Linda~Tuhiwai Smith}.} \bibinfo{year}{2021}\natexlab{}.
\newblock \bibinfo{booktitle}{\emph{Decolonizing methodologies: Research and indigenous peoples}}.
\newblock \bibinfo{publisher}{Bloomsbury Publishing}.
\newblock


\bibitem[Sproat(2010)]%
        {sproat2010language}
\bibfield{author}{\bibinfo{person}{Richard Sproat}.} \bibinfo{year}{2010}\natexlab{}.
\newblock \bibinfo{booktitle}{\emph{Language, technology, and society}}.
\newblock \bibinfo{publisher}{Oxford University Press}.
\newblock


\bibitem[Strubell et~al\mbox{.}(2019)]%
        {strubell2019energy}
\bibfield{author}{\bibinfo{person}{Emma Strubell}, \bibinfo{person}{Ananya Ganesh}, {and} \bibinfo{person}{Andrew McCallum}.} \bibinfo{year}{2019}\natexlab{}.
\newblock \showarticletitle{Energy and Policy Considerations for Deep Learning in NLP}. In \bibinfo{booktitle}{\emph{57th Annual Meeting of the Association for Computational Linguistics (ACL)}}.
\newblock
\urldef\tempurl%
\url{https://doi.org/10.48550/arXiv.1906.02243}
\showURL{%
\tempurl}


\bibitem[Tadei et~al\mbox{.}(2013)]%
        {tadei2013extractive}
\bibfield{author}{\bibinfo{person}{Federico Tadei} {et~al\mbox{.}}} \bibinfo{year}{2013}\natexlab{}.
\newblock \showarticletitle{Extractive institutions and gains from trade: Evidence from colonial Africa}.
\newblock \bibinfo{journal}{\emph{IGIER Working Paper}}  \bibinfo{volume}{536} (\bibinfo{year}{2013}).
\newblock


\bibitem[Theodore et~al\mbox{.}(2023)]%
        {theodore2023maori}
\bibfield{author}{\bibinfo{person}{Reremoana Theodore}, \bibinfo{person}{Amohia Boulton}, {and} \bibinfo{person}{Andrew Sporle}.} \bibinfo{year}{2023}\natexlab{}.
\newblock \showarticletitle{M{\=a}ori Linked Administrative Data: T e Hao Nui-A novel Indigenous Data Infrastructure and Longitudinal Study}.
\newblock \bibinfo{journal}{\emph{The International Indigenous Policy Journal}} \bibinfo{volume}{14}, \bibinfo{number}{1} (\bibinfo{year}{2023}), \bibinfo{pages}{1--16}.
\newblock


\bibitem[Thomas et~al\mbox{.}(2023)]%
        {thomas2023measuring}
\bibfield{author}{\bibinfo{person}{J. Thomas}, \bibinfo{person}{A. McCosker}, \bibinfo{person}{S. Parkinson}, \bibinfo{person}{K. Hegarty}, \bibinfo{person}{D. Featherstone}, \bibinfo{person}{J. Kennedy}, \bibinfo{person}{I. Holcombe-James}, \bibinfo{person}{L. Ormond-Parker}, {and} \bibinfo{person}{L. \&~Ganley}.} \bibinfo{year}{2023}\natexlab{}.
\newblock \bibinfo{title}{Measuring Australia’s Digital Divide: Australian Digital Inclusion Index 2023}.
\newblock
\newblock


\bibitem[Ujomudike(2014)]%
        {Ujomudike2015}
\bibfield{author}{\bibinfo{person}{Philip~Ogochukwu Ujomudike}.} \bibinfo{year}{2014}\natexlab{}.
\newblock \bibinfo{booktitle}{\emph{Ubuntu Ethics}}.
\newblock \bibinfo{publisher}{Springer International Publishing}, \bibinfo{address}{Cham}, \bibinfo{pages}{1--14}.
\newblock
\showISBNx{978-3-319-05544-2}
\urldef\tempurl%
\url{https://doi.org/10.1007/978-3-319-05544-2_428-1}
\showDOI{\tempurl}


\bibitem[Vaswani et~al\mbox{.}(2017)]%
        {vaswani2017attention}
\bibfield{author}{\bibinfo{person}{Ashish Vaswani}, \bibinfo{person}{Noam Shazeer}, \bibinfo{person}{Niki Parmar}, \bibinfo{person}{Jakob Uszkoreit}, \bibinfo{person}{Llion Jones}, \bibinfo{person}{Aidan~N Gomez}, \bibinfo{person}{{\L}ukasz Kaiser}, {and} \bibinfo{person}{Illia Polosukhin}.} \bibinfo{year}{2017}\natexlab{}.
\newblock \showarticletitle{Attention is all you need}.
\newblock \bibinfo{journal}{\emph{Advances in neural information processing systems}}  \bibinfo{volume}{30} (\bibinfo{year}{2017}).
\newblock


\bibitem[Viljoen(2021)]%
        {viljoen2021relational}
\bibfield{author}{\bibinfo{person}{Salome Viljoen}.} \bibinfo{year}{2021}\natexlab{}.
\newblock \showarticletitle{A relational theory of data governance}.
\newblock \bibinfo{journal}{\emph{Yale LJ}}  \bibinfo{volume}{131} (\bibinfo{year}{2021}), \bibinfo{pages}{573}.
\newblock


\bibitem[Wilkins et~al\mbox{.}(2023)]%
        {wilkins2023community}
\bibfield{author}{\bibinfo{person}{Consuelo~H Wilkins}, \bibinfo{person}{Stephania~T Miller}, \bibinfo{person}{Alan~N Richmond}, {and} \bibinfo{person}{Olveen Carrasquillo}.} \bibinfo{year}{2023}\natexlab{}.
\newblock \showarticletitle{Community-Engaged Research-Essential to Addressing Health Inequities}.
\newblock \bibinfo{journal}{\emph{The New England journal of medicine}} \bibinfo{volume}{389}, \bibinfo{number}{21} (\bibinfo{year}{2023}), \bibinfo{pages}{1928--1931}.
\newblock


\bibitem[Zeng et~al\mbox{.}(2023)]%
        {zeng2023english}
\bibfield{author}{\bibinfo{person}{Jie Zeng}, \bibinfo{person}{Ariel~Robert Ponce}, {and} \bibinfo{person}{Yuxin Li}.} \bibinfo{year}{2023}\natexlab{}.
\newblock \showarticletitle{English linguistic neo-imperialism in the era of globalization: A conceptual viewpoint}.
\newblock \bibinfo{journal}{\emph{Frontiers in Psychology}}  \bibinfo{volume}{14} (\bibinfo{year}{2023}), \bibinfo{pages}{1149471}.
\newblock


\bibitem[Zerwas and Dikker(2023)]%
        {wef23}
\bibfield{author}{\bibinfo{person}{Athenia Rodney Michael~Lenihan Zerwas, Felicia} {and} \bibinfo{person}{Suzanne Dikker}.} \bibinfo{year}{2023}\natexlab{}.
\newblock \showarticletitle{Using Data-Driven Community Engagement for Positive Change}.
\newblock \bibinfo{journal}{\emph{World Economic Forum}} (\bibinfo{year}{2023}).
\newblock


\end{thebibliography}

\appendix

\end{document}